\documentclass[11pt,a4paper]{article}

\usepackage[T1]{fontenc}
\usepackage[utf8]{inputenc}
\usepackage{CJKutf8}
\usepackage{times}
\usepackage{marvosym}

\usepackage[margin=2.4cm,headheight=38pt,headsep=18pt,footskip=28pt]{geometry}
\usepackage{setspace}
\usepackage[table,dvipsnames]{xcolor}
\usepackage{colortbl}
\usepackage{graphicx}
\usepackage{booktabs}
\usepackage{array}
\usepackage{multirow}
\usepackage{tabularx}

\usepackage{amsmath,amssymb}

\definecolor{AMapBlue}{HTML}{48B8F0}
\definecolor{AMapDeepBlue}{HTML}{0877C9}
\definecolor{AMapSky}{HTML}{EAF8FF}
\definecolor{AMapLine}{HTML}{9EDBFA}
\definecolor{TextBlack}{HTML}{20242A}

\usepackage[numbers,sort&compress]{natbib}

\usepackage{enumitem}
\setlist{nosep,leftmargin=1.5em}
\usepackage{caption}
\usepackage{fancyhdr}
\usepackage{lastpage}
\usepackage{titlesec}
\usepackage[most]{tcolorbox}
\usepackage{mdframed}
\usepackage[colorlinks=true,linkcolor=AMapDeepBlue,citecolor=AMapDeepBlue,urlcolor=AMapDeepBlue]{hyperref}

\newcommand{\reportname}{FantasyEvo}

\newcommand{\reportlogoheight}{18pt}

\newcommand{\orgname}{AMAP CV Lab \textbar{} Alibaba Group}
\newcommand{\reportcenter}[1]{\raisebox{\dimexpr(\depth-\height)/2\relax}{#1}}
\newlength{\reporttextbaselineoffset}
\newcommand{\reporttextalign}[1]{\raisebox{\reporttextbaselineoffset}{#1}}

\renewcommand{\headrulewidth}{1.2pt}
\renewcommand{\footrulewidth}{0.8pt}
\renewcommand{\headrule}{\hbox to\headwidth{\color{AMapBlue}\leaders\hrule height \headrulewidth\hfill}}
\renewcommand{\footrule}{\hbox to\headwidth{\color{AMapLine}\leaders\hrule height \footrulewidth\hfill}}

\titleformat{\section}
  {\Large\bfseries\color{AMapDeepBlue}}
  {\thesection}{0.75em}{}
  [\vspace{0.32em}{\color{AMapLine}\titlerule[0.6pt]}]
\titleformat{\subsection}
  {\large\bfseries\color{AMapDeepBlue}}
  {\thesubsection}{0.65em}{}
\titleformat{\subsubsection}
  {\normalsize\bfseries\color{AMapDeepBlue}}
  {\thesubsubsection}{0.55em}{}
\titlespacing*{\section}{0pt}{1.55em plus 0.25em minus 0.1em}{0.95em}
\titlespacing*{\subsection}{0pt}{1.25em plus 0.2em minus 0.1em}{0.55em}
\titlespacing*{\subsubsection}{0pt}{1.05em plus 0.15em minus 0.08em}{0.45em}

\newtcolorbox{abstractbox}{
  enhanced,
  breakable,
  colback=AMapSky,
  colframe=AMapBlue,
  coltitle=white,
  colbacktitle=AMapBlue,
  title=Abstract,
  fonttitle=\Large\bfseries,
  halign title=center,
  toptitle=5pt,
  bottomtitle=5pt,
  fontupper=\large,
  boxrule=0.8pt,
  arc=1.5mm,
  left=12pt,
  right=12pt,
  top=10pt,
  bottom=10pt,
  before skip=1.2em,
  after skip=1.4em
}

\newmdenv[
  backgroundcolor=gray!8,
  linewidth=0pt,
  innerleftmargin=10pt,
  innerrightmargin=10pt,
  innertopmargin=5pt,
  innerbottommargin=5pt,
  skipabove=5pt,
  skipbelow=5pt
]{cnblock}

\definecolor{DimAction}{HTML}{0877C9}
\definecolor{DimVisual}{HTML}{2E9E5B}
\definecolor{DimPhysics}{HTML}{E07B00}
\definecolor{DimMemory}{HTML}{8E44AD}
\usepackage{pifont}
\newcommand{\mail}{\textsuperscript{\ding{41}}}
\usepackage{microtype}
\usepackage[capitalize]{cleveref}
\crefname{section}{Sec.}{Secs.}
\Crefname{section}{Section}{Sections}
\Crefname{table}{Table}{Tables}
\crefname{table}{Tab.}{Tabs.}

\definecolor{insightborder}{HTML}{3B28BD}
\newtcolorbox[auto counter]{insight}{
    enhanced,
    colback=black!2,
    colframe=insightborder,
    boxrule=0.55pt,
    arc=2mm,
    boxsep=0pt,
    left=6pt,right=6pt,top=5pt,bottom=5pt,
    before skip=6pt,after skip=6pt,
    fontupper=\normalsize,
    before upper={\textbf{Insight~\thetcbcounter:}\ },
    shadow={0.6mm}{-0.6mm}{0mm}{fill=black!12}
}

\newcommand{\manuscripttitle}{What Transfers in Navigation? Scaling Across Embodiments with Shared and Corrective Experience}
\newcommand{\maketitlesupplementary}{%
  \setcounter{page}{\firstsupplementpage}%
  \suppressfloats[t]%
  \begin{center}
    {\bfseries\color{AMapDeepBlue}\LARGE \manuscripttitle\par}
    \vspace{0.5em}
    {\Large Supplementary Material\par}
    \vspace{1.0em}
  \end{center}
}

\begin{document}
\thispagestyle{fancy}
\renewcommand{\reportname}{NaviBridge}

\begin{center}
  % \vspace*{0.1em}
  {\bfseries\color{AMapDeepBlue}\LARGE \manuscripttitle}

\vspace{0.2em}
{\large 
    Jintao~Kong\textsuperscript{1,2,*},
    Yingnan Guo\textsuperscript{2,*}, 
    Zedong Chu\textsuperscript{1,\mail}, 
    Mu Xu\textsuperscript{1}, 
    Weihuang Chen\textsuperscript{2}, 
    Hongbin Sun\textsuperscript{2}\par
}

\vspace{0.2em}
{\small
  \textsuperscript{*}Equal contribution \quad
  \textsuperscript{\mail}Corresponding author\par
}

\vspace{0.25em}
{\small
  \textsuperscript{1}\,AMAP, Alibaba Group \quad
  \textsuperscript{2}\,School of Artificial Intelligence, Xi'an Jiaotong University\par
}

  \vspace{-0.1em}
  % {\small\color{TextBlack!65}November 2026}
\end{center}

\vspace{-1em}

\begingroup
\let\abstract\abstractbox
\let\endabstract\endabstractbox
\begin{abstract}
To guide data collection across embodiments, we study which navigation behaviors transfer through shared experience and which failures require corrections at states reached after policy deviations.
We propose NaviBridge, a unified world-action model that jointly predicts future observations and actions for indoor and outdoor point-goal navigation, aerial instruction following, and driving.
At fixed source proportions, more data improve outdoor navigation, driving, and aerial instruction following, while indoor performance plateaus earlier.
With the full corpus, NaviBridge achieves 40.74 EPDMS on NAVSIM navhard and 55.43\% success on OpenFly.
Behavior analysis links shared training to safer driving, more consistent replanning, and more reliable aerial task completion after goal approach, highlighting different demands for stopping and continued road progress.
Corrective training achieves 100\% indoor and 95.11\% outdoor success on ABotN-PointBench, exceeding the larger ABot-N1 trained on more data.
We deploy NaviBridge on a quadruped and a UAV and make open-loop predictions on real driving recordings.
\end{abstract}

\endgroup

\noindent\begin{minipage}{\textwidth}
    \centering
    \captionsetup{type=figure}
    \includegraphics[width=\linewidth]{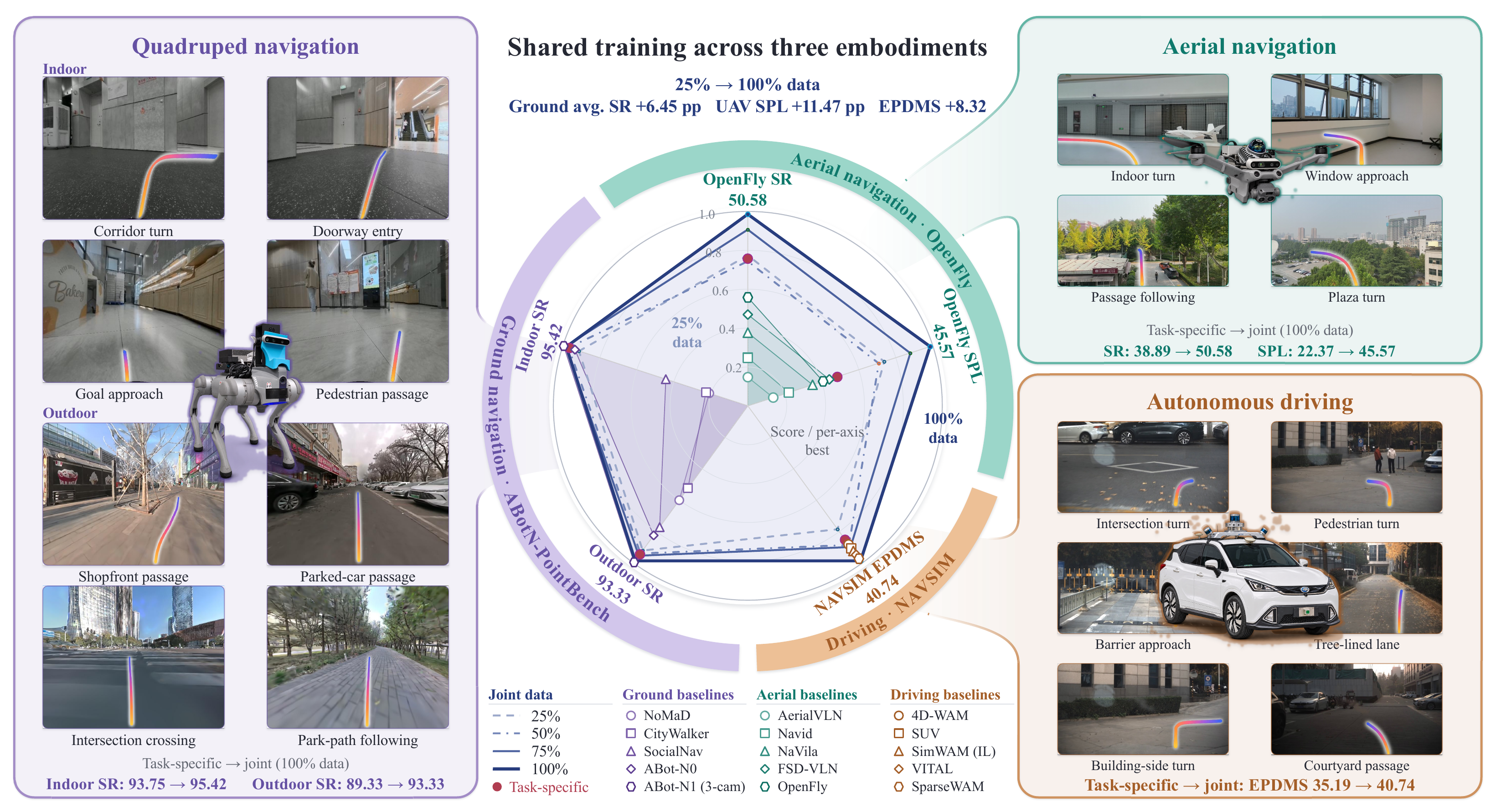}
    \caption{NaviBridge across ground navigation, aerial instruction following, and autonomous driving. The radar chart compares joint training at four data scales with task-specific models and published baselines. The surrounding panels show quadruped and UAV deployments and open-loop predictions on real driving recordings.}
    \label{fig:overview}
\end{minipage}\par
\vspace{0.5\baselineskip}%

\Needspace{6\baselineskip}
\section{Introduction}
\label{sec:intro}

Navigation in new environments or on new platforms requires demonstrations suited to the robot's observations and motion, yet data availability varies across tasks.
Ground robots, UAVs, and vehicles make many of the same navigation decisions, including choosing traversable paths, avoiding obstacles, and approaching goals.
A unified policy can learn these common behaviors from readily available demonstrations, allowing new data collection to focus on the target platform's specific motion and task requirements~\cite{shah2023gnm,yang2024cross}.

\textbf{What transfers in navigation?} Suomela et al.\ show that broader geographic coverage improves generalization in ground navigation, with diminishing returns from more data at the same locations~\cite{suomela2026navigation}.
Ego4WAM distinguishes data duration from task diversity and reports gains and losses from long-tail expansion across manipulation capabilities and tasks~\cite{sun2026ego4wam}.
Together with evidence on environment and object diversity in manipulation~\cite{lin2025datascaling}, these studies show that data composition shapes scaling gains.
However, task scores leave open which navigation behaviors transfer and which states require supervision from policy rollouts.
We trace changes in scores to execution behavior to guide data collection.

Our experiments reveal two patterns.
First, gains vary by task. Joint indoor and outdoor training improves both ground tasks over separate training. The four-source model outperforms the corresponding single-task models on aerial instruction following and driving, while ground scores are slightly lower than with joint indoor and outdoor training.
At fixed source proportions, more data improve outdoor navigation, driving, and aerial instruction following, while indoor performance plateaus earlier.
Second, progress toward a goal can still end in failure. A UAV may approach a goal and leave again, and a ground policy may take a wrong turn, overshoot a junction, or collide while recovering.

These observations motivate two complementary forms of experience.
\textbf{Shared experience} uses traversal, obstacle avoidance, and goal-approach demonstrations across tasks. Its value depends on whether those behaviors match the target task's needs at each stage.
\textbf{Corrective experience} supplies recovery actions from positions and headings the policy visits after deviating, which routine demonstrations rarely cover.
Together, they expand coverage of target behaviors and policy-visited states.

To study these two forms of experience, we develop NaviBridge, a unified world-action model for indoor and outdoor point-goal navigation, aerial instruction following, and driving.
It jointly predicts future observations and actions in a common representation~\cite{nvidia2025cosmospredict2,kim2026cosmospolicy}.
We vary data volume and source composition separately. Driving metrics and UAV trajectories link shared training to safer driving, better drivable-area compliance, more consistent replanning, and more reliable aerial task completion after goal approach.
These gains align with detours and terminal control in point-goal data, while changes in driving progress highlight the role of the execution stage.
Starting from the joint indoor and outdoor policy, we use DAgger to test whether expert recovery actions at states visited by the policy improve safe completion once routine performance plateaus~\cite{ross2011dagger}.

These results suggest two guidelines for data collection. \textbf{For data-scarce tasks, use readily available auxiliary data that match target behaviors.}
Choose auxiliary sources that fill gaps in traversal, obstacle avoidance, or goal approach, and use each task's scaling curve to guide further collection.

\textbf{When routine performance plateaus, collect corrective demonstrations from closed-loop failures.}
With the 2B backbone held fixed, adding 83.6k indoor correction samples raises indoor success from 96.67\% to 100\% and improves outdoor success.
The corrected policy also surpasses the larger ABot-N1 in both environments~\cite{gong2026abotn1}.

Our main contributions are as follows.
\begin{itemize}
    \item \textbf{Cross-embodiment scaling.} With NaviBridge's common representation, we examine scaling at fixed source proportions, changes in source composition, and corrections at policy-visited states.
    \item \textbf{Behavior analysis.} Driving metrics, paired UAV trajectories, and recovery cases connect gains to traversal, terminal control, and recovery, guiding shared and corrective data collection.
    \item \textbf{Benchmarks and real-world evaluation.}  NaviBridge achieves state-of-the-art results in the reported comparisons on ABotN-PointBench, OpenFly, and single-front-view NAVSIM navhard. We demonstrate closed-loop quadruped and UAV deployments and open-loop trajectory prediction on real driving recordings.
\end{itemize}

\section{Related Work}
\label{sec:related}

\textbf{Data scaling and cross-embodiment learning.}
DD-PPO scales point-goal learning to billions of simulated frames in Habitat~\cite{savva2019habitat,wijmans2020ddppo}.
GNM and ViNT learn visual navigation from demonstrations collected by different robots~\cite{shah2023gnm,shah2023vint}.
Open X-Embodiment demonstrates cross-robot transfer in manipulation~\cite{openx2024}, and Yang et al.\ study manipulation and navigation with goal-reaching policies~\cite{yang2024cross}.
Scaling experiments vary geographic coverage, object and environment diversity, and the range of manipulation tasks~\cite{suomela2026navigation,lin2025datascaling,sun2026ego4wam}.
We examine how data volume and source composition affect specific navigation behaviors.

\textbf{Behavior coverage and data composition.}
DAgger queries experts at states visited by the learned policy~\cite{ross2011dagger}.
DART injects noise during demonstrations, and ChauffeurNet synthesizes perturbed driving trajectories to teach recovery~\cite{laskey2017dart,bansal2019chauffeurnet}.
We examine how corrective experience complements the traversal and terminal behaviors learned from shared data.

\textbf{World-action modeling.}
The Unified World Models framework couples video and action diffusion to learn from robot trajectories and action-free video~\cite{zhu2025uwm}.
Sparse-WAM reduces inference cost by selecting action-relevant future-frame tokens~\cite{xie2026sparse}.

\textbf{Navigation evaluation and behavior analysis.}
NAVSIM and subsequent work on pseudo-simulation evaluate safety, road compliance, progress, and comfort in driving~\cite{dauner2024navsim,cao2025pseudo,navsim2026metrics}.
AerialVLN extends vision-language navigation to UAVs~\cite{liu2023aerialvln}, and OpenFly evaluates goal reaching and task completion across seen and unseen scenes~\cite{gao2026openfly}.
We compare predictions and rollouts on matched tasks to identify the execution stages behind score changes.

\section{Method}
\label{sec:method}

NaviBridge is a unified world-action model that learns from shared experience across environments and embodiments, together with corrective supervision at policy-visited states.
We describe the representation, training objective, and construction of these two forms of experience.

\subsection{Visual and Action Representation}
\label{sec:representation}

NaviBridge uses a video diffusion model to jointly predict future visual observations and actions conditioned on visual history and a task specification.
Each training sample contains four past images, the current image, five future waypoints, and the corresponding future images.
Indoor and outdoor tasks specify a local goal position, aerial tasks provide a natural-language instruction, and driving tasks provide a textual navigation command.
This shared interface lets the policy learn how visual changes relate to motion across tasks.

We represent actions as two-dimensional displacements in the current egocentric frame, with positive $x$ to the right and positive $y$ forward.
For source $s$, we normalize the difference between successive waypoints by its motion scale $\kappa_s$:
\begin{equation}
    a_k = \frac{p_k-p_{k-1}}{\kappa_s},\qquad
    p_0=\mathbf{0},\quad k=1,\ldots,5.
    \label{eq:actions}
\end{equation}
Here $p_k$ is the position of waypoint $k$ in meters relative to the current observation, and $a_k$ is its normalized increment.
During sample construction, source-specific normalization maps motions of different magnitudes to a shared action space.
At execution, we rescale and accumulate predicted increments to recover waypoints; aerial samples represent motion in the horizontal plane.

\subsection{Conditional World-Action Modeling}
\label{sec:modeling}

NaviBridge uses the Cosmos-Predict2-2B diffusion backbone and follows the latent-space formulation of Cosmos Policy~\cite{nvidia2025cosmospredict2,kim2026cosmospolicy}.
The visual encoder maps images to latent variables, and future actions occupy dedicated latent slots.
Given observation history and task conditioning, the diffusion Transformer jointly models future actions and visual latents:
\begin{equation}
    p_\theta\!\left(a_{1:H},z_{1:H}\mid o_{-K:0},c\right),
    \qquad K=4,\quad H=5.
    \label{eq:conditional}
\end{equation}
Here $\theta$ denotes the model parameters, $o$ denotes image observations, $a$ denotes actions, $z$ denotes image latents, and $c$ denotes the task condition.
The model receives $K$ past observations and the current image as context, with goal or text encoding supplying the task condition.
The denoising target combines actions and visual latents over the $H$ future steps, linking trajectory prediction to the expected visual changes.

The training objective weights the diffusion loss by data source:
\begin{equation}
    \mathcal L(\theta)=\sum_{s\in\mathcal S}\lambda_s
    \mathbb E_{x\sim\mathcal D_s,\,\sigma,\,\epsilon}
    \bigl[\ell_{\mathrm{EDM}}(\theta;x,\sigma,\epsilon)\bigr].
    \label{eq:training}
\end{equation}
For source $s$ in $\mathcal S$, $\mathcal D_s$ defines the training distribution and the sampling proportion implements weight $\lambda_s$.
A sample $x$ contains the observation history, task condition, future actions, and future images.
The noise level $\sigma$ and Gaussian noise $\epsilon$ enter the weighted denoising loss $\ell_{\mathrm{EDM}}$.
Both the action and future-image slots provide supervision for the diffusion Transformer. We keep the visual and text encoders frozen during training.

\subsection{Shared and Corrective Experience}
\label{sec:data}

Shared experience comprises ground planning trajectories, aerial instruction trajectories, and driving recordings expressed through the same observation and action interface.
Corrective experience pairs observations from ground policy rollouts with expert recovery actions.
Supplementary Secs.~A and F detail the training recipes and behavior coverage.

\textbf{Indoor and outdoor data.}
We plan and smooth indoor trajectories in InteriorGS scenes and render sequential observations along them~\cite{miao2026towards}.
Outdoor data come from real scenes scanned and reconstructed with the ABot-N platform \cite{gong2026abotn1} and cover turning, obstacle avoidance, and crossing trajectories.
We extract training windows centered on the current observation, yielding about 599k indoor and 6.09M outdoor samples.
The resulting datasets differ in visual scale, traversable space, and motion distributions, providing a basis for studying cross-environment transfer.

\textbf{Aerial and driving data.}
We resample discrete OpenFly trajectories into continuous horizontal waypoints paired with natural-language instructions and front-view images~\cite{gao2026openfly}.
The driving samples pair NAVSIM ego trajectories in the current vehicle frame with the corresponding recorded images.
The aerial and driving corpora contain about 1.86M and 53.1k samples, respectively, and supervise aerial instruction following and road motion.

\textbf{Policy correction data.}
We use the same procedure indoors and outdoors. We collect deviation and recovery segments during policy rollouts and use A* paths to supervise recovery from the states the policy visits.
The resulting observations and recovery actions form about 83.6k indoor and 199.7k outdoor samples.
We mix these corrections with base navigation data to teach recovery alongside routine navigation.

\AddToHook{env/tabular*/begin}[navibridge]{\setlength{\tabcolsep}{.65\tabcolsep}}
\section{Data Scaling and Behavioral Transfer}
\label{sec:experiments}

In this section, we vary the amount and sources of shared experience to measure which navigation tasks benefit from data expansion. Paired behavior analyses locate the gains and failures during execution. Corrective training then tests whether supervision at states reached after deviations improves recovery.

\subsection{Experimental Setup}
\label{sec:setup}

\paragraph{Tasks and metrics.}
For ground point-goal navigation, we report success rate (SR) and success weighted by path length (SPL) on 240 indoor and 225 outdoor ABotN-PointBench tasks~\cite{gong2026abotn1}.
Both metrics use the benchmark's arrival and collision criteria for each task.
For driving, we report EPDMS and its components on NAVSIM navhard V2 with reactive traffic and a single front-facing camera.
For aerial instruction following, we report SR and SPL on the publicly released OpenFly Test Seen and Test Unseen scenes~\cite{gao2026openfly}.
The model predicts horizontal waypoints, and the simulator takes flight altitude from the reference trajectory.
Supplementary Sec.~B describes the evaluation protocols.

\textbf{Training comparisons.}
We fix the four sources and their mixture weights and compare training on 25\%, 50\%, 75\%, and 100\% of the available data.
To assess source composition, we compare single-domain models, the joint indoor and outdoor model, and the four-source NaviBridge model. NaviBridge uses the same weights across tasks.
The correction models start from the joint ground policy and mix base data with indoor, outdoor, or joint corrections.
Supplementary Secs.~A and C give the training recipes, budgets, checkpoint selection, and complete scaling results.

\textbf{Behavior analysis.}
We compare ground collisions and path efficiency on matched tasks and use DAgger to analyze recovery.
Driving analysis covers safety, planning consistency, and road progress, with paired observations for safety pass rates and predicted path length.
For aerial instruction following, we pair episodes by task and analyze goal-region entry, the final position, and predicted displacement at stopping.
Supplementary Secs.~D, E, and G detail the paired analyses.

\subsection{Scaling at Fixed Source Composition}
\label{sec:scaling}

\begin{figure*}[t]
    \centering
    \includegraphics[width=.93\textwidth]{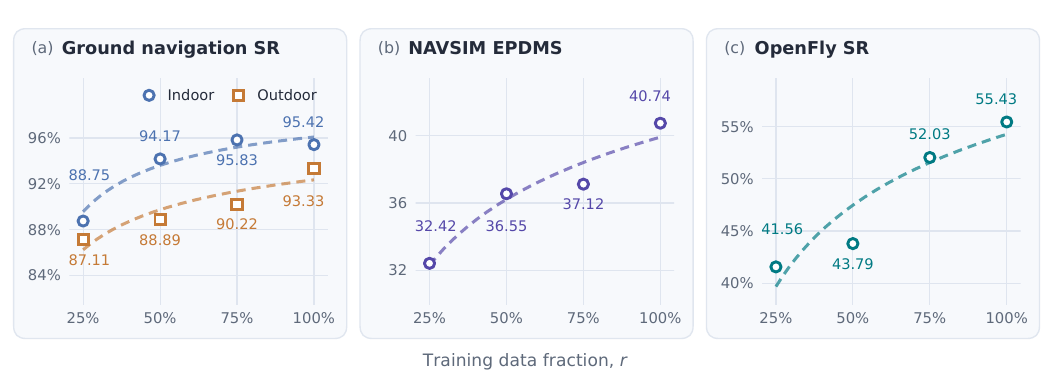}
    \caption{Performance versus the fraction of available training data at fixed source composition. (a) Indoor and outdoor point-goal SR. (b) NAVSIM EPDMS. (c) OpenFly SR. Open markers are measurements, and dashed lines are power-law fits. Supplementary Sec.~C gives the complete scores and fitting method.}
    \label{fig:scaling}
\end{figure*}

Prior work shows that geographic coverage and task diversity shape the gains from adding training data~\cite{suomela2026navigation,sun2026ego4wam}.
To compare tasks at fixed composition, we vary the amount of data available from each of the four sources by the same fraction.
The full-data NaviBridge model gives the 100\% endpoint in \cref{fig:scaling}, and Supplementary Table~S2 reports the complete results.

The scaling curves show gains across tasks, with different rates of improvement as data volume increases.
From 25\% to the full corpus, outdoor SR rises from 87.11\% to 93.33\%, and driving EPDMS rises from 32.42 to 40.74.
OpenFly SR rises from 41.56\% to 55.43\%, an increase of 249 successful tasks.
The main indoor gain occurs between 25\% and 50\%, when SR rises from 88.75\% to 94.17\%, before leveling off.

\begin{insight}
Outdoor navigation, driving, and aerial instruction following continue to improve with more data, while routine indoor navigation saturates earlier.
\end{insight}

\subsection{Source Composition and Task Coverage}
\label{sec:sources}

Adding sources changes both environment coverage and the supervision available to each task.
We compare task-specific, joint ground, and four-source models on matched evaluation tasks to measure how performance changes as coverage expands.

\begin{table}[t]
    \caption{Ground navigation under different source compositions (\%). I+O denotes joint indoor and outdoor training, and NaviBridge uses all four sources. Means weight unrounded domain scores equally, and bold marks the best score in each column.}
    \label{tab:ground-sources}
    \centering
    \small
    \setlength{\tabcolsep}{1.8pt}
    \begin{tabular*}{\columnwidth}{@{\extracolsep{\fill}}lcccccc@{}}
        \toprule
        Model & \multicolumn{2}{c}{Indoor} & \multicolumn{2}{c}{Outdoor} & \multicolumn{2}{c}{Mean} \\
        \cmidrule(lr){2-3}\cmidrule(lr){4-5}\cmidrule(l){6-7}
        & SR & SPL & SR & SPL & SR & SPL \\
        \midrule
        Indoor only & 93.75 & 90.36 & 68.00 & 61.90 & 80.875 & 76.133 \\
        Outdoor only & 23.75 & 23.75 & 89.33 & 88.46 & 56.542 & 56.107 \\
        Joint ground$_{\mathrm{I+O}}$ & \textbf{96.67} & \textbf{92.22} & \textbf{93.78} & \textbf{92.53} & \textbf{95.222} & \textbf{92.377} \\
        NaviBridge & 95.42 & 91.90 & 93.33 & 92.32 & 94.375 & 92.110 \\
        \bottomrule
    \end{tabular*}
\end{table}

\textbf{Cross-environment transfer.}
The joint ground model achieves 96.67\% indoor and 93.78\% outdoor SR, compared with 93.75\% and 89.33\% for the respective single-domain models (\cref{tab:ground-sources}).
Specialists transfer poorly across domains, while the joint policy serves both.
The indoor specialist and joint policy reach every indoor goal, with fewer collisions accounting for most gains on harder tasks.

\textbf{Cross-embodiment expansion.}
NaviBridge samples outdoor, indoor, UAV, and driving data with weights of 0.40, 0.35, 0.20, and 0.05, respectively.
Its indoor and outdoor SRs are 95.42\% and 93.33\%, slightly below those of the joint ground model.
Relative to single-task training, driving EPDMS increases from 35.19 to 40.74, and OpenFly SR increases from 41.97\% to 55.46\% on the common tasks (\cref{tab:multitask-sources}).

\begin{insight}
Joint indoor and outdoor training provides a strong base for point-goal navigation.
The four-source mixture serves driving and aerial instruction following with only a small drop in ground navigation scores.
\end{insight}

\begin{table}[ht]
    \caption{Driving and aerial instruction following after separate or joint training. NaviBridge uses the same weights for both tasks.}
    \label{tab:multitask-sources}
    \centering
    \small
    \setlength{\tabcolsep}{4pt}
    \begin{tabularx}{\columnwidth}{@{}Xccc@{}}
        \toprule
        Training & NAVSIM & \multicolumn{2}{c}{OpenFly} \\
        \cmidrule(lr){2-2}\cmidrule(l){3-4}
                 & EPDMS & SR (\%) & SPL (\%) \\
        \midrule
        Task-specific & 35.19 & 41.97 & 23.87 \\
        NaviBridge & \textbf{40.74} & \textbf{55.46} & \textbf{49.26} \\
        \bottomrule
    \end{tabularx}
\end{table}

\subsection{What Transfers, and What Needs Correction?}
\label{sec:behavior}

We connect driving safety and progress, and aerial goal approach and stopping, to the behaviors supervised by the training data.
These analyses identify how shared experience matches each task before we examine corrective experience for recovery after deviations.

\subsubsection{Driving Safety, Planning Consistency, and Progress}
\label{sec:driving-behavior}

On NAVSIM navhard, we compare NaviBridge with a model trained on the same driving corpus alone.
We ask which driving behaviors benefit from shared training and which ground navigation experiences correspond to these changes.

\textbf{Behavior changes.}
The main gains are in safety and replanning consistency.
In Stage 2, no at-fault collision, drivable-area compliance, and time-to-collision scores improve by 1.23, 2.31, and 2.34 points, respectively.
On 5,462 common Stage 2 observations, predictions satisfying all four safety and compliance constraints rise from 58.35\% to 62.05\%.
The constraints require no at-fault collisions and compliance with drivable areas, driving directions, and traffic lights.
The gain is 3.70 percentage points, with a scenario-group bootstrap 95\% confidence interval of [2.45, 4.99].
Extended Comfort, which measures consistency between successive plans, improves by 22.67 and 14.65 points in the two stages, while Stage 2 progress falls by 3.04 points.
On the 2,988 paired Stage 2 observations where both models satisfy all four safety and compliance constraints, NaviBridge's predicted four-second paths are 0.22\,m shorter on average.
Supplementary Sec.~E gives the full metrics and predicted-path analysis.

\textbf{Sources of experience.}
These changes align with traversal and terminal supervision in the auxiliary data.
In the outdoor corpus, 39.48\% of windows come from dedicated obstacle-avoidance collection.
In the standard indoor category, 54.02\% come from trajectories at least 20\% longer than the straight-line distance between their endpoints.
These demonstrations connect obstacles to traversable paths, supplying spatial experience that can transfer to driving.
Small terminal motions in point-goal data may also shorten some driving predictions.
Supplementary Secs.~F.2 and F.3 report the terminal-window and obstacle-avoidance statistics.

\subsubsection{Aerial Goal Approach and Stopping}
\label{sec:aerial-behavior}

Aerial instruction following requires finishing at a designated location, whereas driving typically requires continued road progress.
We pair complete OpenFly rollouts from the two models by task in the Seen and Unseen scenes.
We count goal approach when any recorded position after the start lies within 20\,m of the goal in three dimensions.
We count completion when the endpoint lies within the same region.

The joint model improves SR in all eight scenes and records 242 net additional successes on common tasks.
The gain comprises 48 more rollouts entering the goal region and 194 fewer rollouts ending outside after entry.
Fewer failures after entry account for roughly four-fifths of the net gain.
In GZ, the joint model enters the goal region on fewer tasks while completing more tasks overall. This illustrates why we examine approach and completion separately.

NaviBridge uses continuous waypoint control and signals a stop when the waypoints selected for execution imply a net horizontal displacement below 1\,m.
On 1,046 tasks where both models enter the goal region and share a waypoint stopping rule, the stop rate rises from 54.59\% to 72.66\%.
Most added stops occur after the first entry, as the joint model more often predicts a stopping action near the goal.
Supplementary Sec.~G.3 reports the paired stopping analysis.

As a point-goal trajectory approaches its destination, future waypoints cluster around the goal.
Terminal padding supplies further supervision from short motions down to zero displacement.
Indoor and outdoor point-goal data account for about three-quarters of windows with repeated terminal waypoints under the joint sampling distribution.
This experience aligns with the need to finish at a designated location in OpenFly and with higher completion after goal approach.

\begin{insight}
Terminal supervision matches the need to stop at the goal in aerial instruction following and may shorten predicted paths in driving. Transfer depends on what the target task requires during execution.
\end{insight}

\subsubsection{Corrective Experience: Recovery after Deviations}
\label{sec:correction}

Once the joint ground model handles routine traversal, adding aerial and driving sources offers little further benefit for point-goal navigation (\cref{tab:ground-sources}).
Recovery requires acting from positions and headings that routine expert trajectories rarely visit, such as those reached after entering the wrong corridor or overshooting a junction.
We therefore collect rollouts from the joint ground policy and provide expert recovery actions at the states it visits.
\Cref{tab:correction} compares indoor, outdoor, and joint correction recipes by evaluating one checkpoint per recipe in both domains.

\begin{table}[t]
    \caption{Point-goal performance under three correction recipes (\%). Means weight unrounded domain scores equally, and bold marks the best score in each column.}
    \label{tab:correction}
    \centering
    \small
    \setlength{\tabcolsep}{1.8pt}
    \begin{tabular*}{\columnwidth}{@{\extracolsep{\fill}}lcccccc@{}}
        \toprule
        Training & \multicolumn{2}{c}{Indoor} & \multicolumn{2}{c}{Outdoor} & \multicolumn{2}{c}{Mean} \\
        \cmidrule(lr){2-3}\cmidrule(lr){4-5}\cmidrule(l){6-7}
        & SR & SPL & SR & SPL & SR & SPL \\
        \midrule
        Joint ground & 96.67 & 92.22 & 93.78 & 92.53 & 95.222 & 92.377 \\
        Indoor DAgger & \textbf{100.00} & \textbf{95.76} & 94.67 & 94.13 & \textbf{97.333} & \textbf{94.945} \\
        Outdoor DAgger & 95.83 & 94.33 & \textbf{95.11} & \textbf{94.99} & 95.472 & 94.659 \\
        Joint DAgger & \textbf{100.00} & 95.56 & 94.22 & 93.81 & 97.111 & 94.688 \\
        \bottomrule
    \end{tabular*}
\end{table}

All three recipes improve mean SPL over the joint ground baseline.
Indoor correction obtains the highest mean scores.
Outdoor correction performs best outdoors, while joint correction succeeds on every indoor task and maintains high outdoor performance.

Indoor correction shows that a small set of recovery demonstrations can benefit both environments.
The recipe uses the same policy architecture and adds 83.6k corrective samples, equivalent to 1.25\% of the base point-goal corpus.
Indoor SR and SPL improve from 96.67\%/92.22\% to 100\%/95.76\%. Outdoors, they improve from 93.78\%/92.53\% to 94.67\%/94.13\%.

\begin{figure}[t]
    \centering
    \includegraphics[width=\columnwidth]{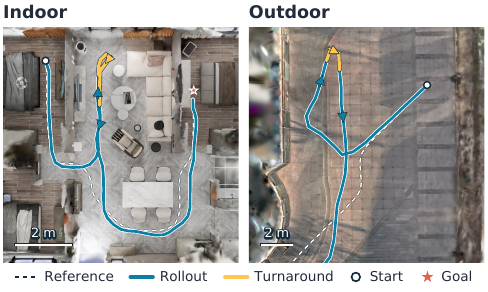}
    \caption{Top-down trajectory views after indoor and outdoor DAgger training.}
    \label{fig:recovery}
\end{figure}

\Cref{fig:recovery} illustrates route recovery after deviations.
In the indoor example, the joint ground model reaches the goal with collisions. The corrected policy turns back after deviating, rejoins the route, and reaches the goal collision-free.
In the outdoor example, the rollout deviates several meters before returning to the junction and continuing toward the goal.

\begin{insight}
Corrective experience improves recovery and navigation performance at a fixed policy capacity, with benefits that can extend beyond the environment where corrections were collected.
\end{insight}

\section{From Transfer Insights to Benchmark Performance}
\label{sec:benchmarks}

Shared training improves driving and aerial instruction following, and corrections improve recovery in ground navigation (\cref{sec:experiments}).
We therefore benchmark the same four-source model on driving and aerial instruction following and compare the indoor, outdoor, and joint correction variants of the ground policy.
\Cref{tab:benchmark-summary} compares these models with published methods across the three benchmarks.

\begin{table*}[t]
    \caption{Comparison across three navigation benchmarks. SR and SPL are percentages; bold marks the best score in each column.
    (a) Subscripts I, O, and I+O denote indoor, outdoor, and joint correction of the ground policy. ABot-N1 uses three camera views~\cite{gong2026abotn1}.
    (b) Single-front-view methods on navhard V2 with reactive traffic.
    (c) Published OpenFly results. NaviBridge counts tasks with changes in reference altitude as failures on the released scenes.}
    \label{tab:benchmark-summary}
    \label{tab:openfly-benchmark}
    \centering
    \small
    \setlength{\tabcolsep}{2pt}
    \begin{minipage}[t]{.365\textwidth}
        \centering
        \textbf{(a) ABotN-PointBench}\\[3pt]
        \begin{tabular*}{\linewidth}{@{\extracolsep{\fill}}lrrrr@{}}
            \toprule
            Method & \multicolumn{2}{c}{Indoor} & \multicolumn{2}{c}{Outdoor} \\
            \cmidrule(lr){2-3}\cmidrule(l){4-5}
            & SR & SPL & SR & SPL \\
            \midrule
            ABot-N0\cite{chu2026abotn0} & 89.6 & 85.6 & 76.9 & 76.9 \\
            ABot-N1\cite{gong2026abotn1} & 95.4 & 93.7 & 92.9 & 91.4 \\
            \midrule
            NaviBridge$_{\mathrm I}$ & \textbf{100.00} & \textbf{95.76} & 94.67 & 94.13 \\
            NaviBridge$_{\mathrm O}$ & 95.83 & 94.33 & \textbf{95.11} & \textbf{94.99} \\
            NaviBridge$_{\mathrm{I+O}}$ & \textbf{100.00} & 95.56 & 94.22 & 93.81 \\
            \bottomrule
        \end{tabular*}
    \end{minipage}\hfill
    \begin{minipage}[t]{.245\textwidth}
        \centering
        \textbf{(b) NAVSIM navhard}\\[3pt]
        \begin{tabular*}{\linewidth}{@{\extracolsep{\fill}}lr@{}}
            \toprule
            Method & Score \\
            \cmidrule(l){2-2}
                   & EPDMS \\
            \midrule
            SUV\cite{yuan2026suv} & 36.9 \\
            SimWAM (IL)\cite{zhao2026simwam} & 37.6 \\
            VITAL\cite{liang2026vital} & 38.6 \\
            SparseWAM\cite{yang2026sparsewam} & 39.6 \\
            \midrule
            NaviBridge & \textbf{40.74} \\
            \bottomrule
        \end{tabular*}
    \end{minipage}\hfill
    \begin{minipage}[t]{.355\textwidth}
        \centering
        \textbf{(c) OpenFly}\\[3pt]
        \begin{tabular*}{\linewidth}{@{\extracolsep{\fill}}lrrrr@{}}
            \toprule
            Method & \multicolumn{2}{c}{Seen} & \multicolumn{2}{c}{Unseen} \\
            \cmidrule(lr){2-3}\cmidrule(l){4-5}
            & SR & SPL & SR & SPL \\
            \midrule
            Navid\cite{zhang2024navid} & 13.0 & 11.6 & 10.8 & 5.0 \\
            NaVila\cite{cheng2025navila} & 20.3 & 17.8 & 14.7 & 9.6 \\
            FSD-VLN\cite{zhu2026fsdvln} & 26.7 & 22.8 & 13.6 & 10.7 \\
            OpenFly\cite{gao2026openfly} & 28.8 & 18.7 & 26.2 & 18.0 \\
            \midrule
            NaviBridge & \textbf{51.55} & \textbf{46.43} & \textbf{47.28} & \textbf{42.59} \\
            \bottomrule
        \end{tabular*}
    \end{minipage}
\end{table*}

\subsection{Ground Point-Goal Navigation}
\label{sec:ground-benchmark}

\textbf{ABotN-PointBench.}
\Cref{tab:benchmark-summary}(a) compares the correction configurations with ABot-N0 and ABot-N1.
The indoor correction model achieves 100\% SR and 95.76\% SPL, exceeding ABot-N1 by 4.60 and 2.06 percentage points, respectively.
The outdoor correction model achieves 95.11\% SR and 94.99\% SPL, gains of 2.21 and 3.59 percentage points.
These are the highest respective-domain scores in the comparison.
Joint correction achieves 100\% indoor and 94.22\% outdoor SR with one checkpoint, also exceeding ABot-N1 in both domains.

\textbf{Training data and policy scale.}
ABot-N1 pairs a 4B reasoner with a 2B action expert and uses about 30M samples across its full multitask navigation pretraining corpus~\cite{gong2026abotn1}.
NaviBridge uses a 2B diffusion policy backbone and about 8.61M samples across four sources.
The three ground correction recipes use 6.78M to 6.98M base and corrective samples in total.
Joint correction exceeds ABot-N1's 95.4\% indoor and 92.9\% outdoor SR.
Targeted correction further improves indoor SPL and outdoor SR and SPL.
These results emphasize behavior and state coverage alongside model capacity and architecture.
Supervision for complementary behaviors and recovery states improves performance while retaining the same policy backbone.

\subsection{Autonomous Driving}
\label{sec:driving-benchmark}

\textbf{NAVSIM.}
The four-source NaviBridge model scores 40.74 EPDMS on navhard, exceeding SparseWAM's 39.6 in the single-front-view comparison (\cref{tab:benchmark-summary}(b)).
It uses front-view image history and a navigation command, with two denoising steps and one predicted trajectory.
Published driving models span different input and inference settings, including additional cameras or vehicle-state inputs~\cite{xiong2026terratransfer,chen2026fpddrive,xu2026toad,hong2026drivefuture}.
Supplementary Sec.~B.3 specifies our five-waypoint time mapping.

\subsection{Aerial Instruction Following}
\label{sec:aerial-benchmark}

\textbf{OpenFly.}
We evaluate the same NaviBridge checkpoint used for driving on the released Test Seen and Test Unseen scenes.
In \cref{tab:benchmark-summary}(c), we count all tasks with changes in reference altitude as failures for NaviBridge to ensure a fair comparison.

NaviBridge obtains the highest SR and SPL in the table on both splits.
Relative to OpenFly, SR/SPL improve by 22.75/27.73 percentage points on Seen and 21.08/24.59 on Unseen.
Supplementary Secs.~B.2 and G.1 detail the scoring protocol and baseline score aggregation.

\section{Real-World Experiments}
\label{sec:real-world}

\begin{figure*}[t]
    \centering
    \includegraphics[width=\textwidth,height=\textheight,keepaspectratio]{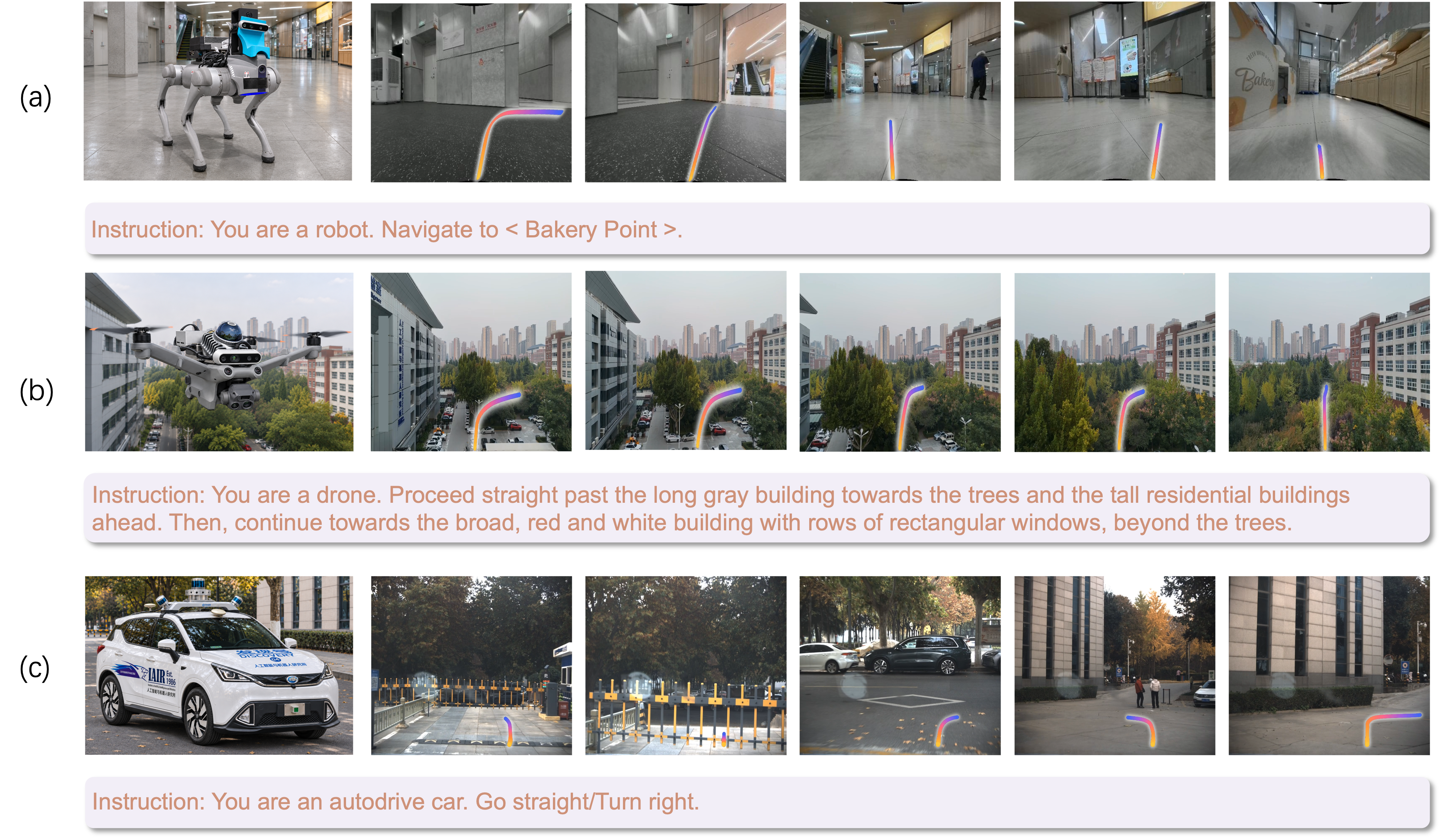}
    \caption{Real-world trajectory visualizations for a quadruped (a), a UAV (b), and autonomous driving (c). The quadruped and UAV run in closed loop. Driving uses open-loop predictions under local road regulations.}
    \label{fig:real-world}
\end{figure*}

The quadruped tests point-goal navigation without real-world fine-tuning, and the UAV tests instruction following after few-shot adaptation.
Both run closed loop with one denoising step and execute predicted waypoints through their controllers.
Under local road-use restrictions, we use a NAVSIM-style open-loop evaluation and predict trajectories from real-road recordings with two denoising steps.
\Cref{fig:real-world} shows representative trajectories across all three embodiments and their environments.

\subsection{Indoor Quadruped Navigation}
\label{sec:quadruped}

The quadruped executes indoor point-goal navigation on a Jetson Orin without real-world fine-tuning, with about 800\,ms per inference.
In \cref{fig:real-world}(a), the robot starts facing a wall, turns into an open corridor, passes an escalator and a display area, and follows the counter to the bakery.
It adapts its predicted waypoints to the changing views and completes turns, detours, and goal approach from a low camera viewpoint.

\subsection{Few-Shot Aerial Instruction Following}
\label{sec:uav-real}

The UAV follows natural-language instructions involving landmarks, directions, and ordered turns on three outdoor and two indoor tasks.
We fine-tune a single model jointly on four complete demonstrations, one from each of four tasks.
We evaluate those four tasks in the one-shot setting at altitudes different from those used to collect the demonstrations. A held-out outdoor task uses the same adapted model for task-level zero-shot evaluation.

The UAV communicates over a local network with a computer equipped with an RTX 3070. The computer takes about 1\,s to predict horizontal waypoints, which the flight controller executes.
In \cref{fig:real-world}(b), the UAV follows an outdoor passage, turns near buildings as instructed, and reaches the endpoint.
Supplementary Fig.~S5 shows the other indoor and outdoor tasks.

\subsection{Open-Loop Prediction on Real Roads}
\label{sec:driving-real}

NaviBridge takes four historical front-view images, the current image, and a navigation command to predict future waypoints from a driving recording.
The route covers a plaza in front of a building, roads lined with parked vehicles, a tree-lined segment, and narrow passages between buildings.
The commands include going straight and turning left or right.

\Cref{fig:real-world}(c) includes a parking barrier absent from the driving corpus.
In the recorded sequence, the vehicle approaches the closed barrier, waits for it to lift, and proceeds. The predicted path shortens as the vehicle approaches and extends again after the barrier opens.
On an unstructured roadway, the predicted path also leaves clearance for pedestrians.
The driving data contain neither scenario, while the outdoor point-goal corpus includes analogous behaviors.
These examples illustrate the sharing of navigation experience across embodiments.

\section{Conclusion}
\label{sec:conclusion}

We investigated what transfers across navigation environments and embodiments through shared and corrective experience in NaviBridge.
Scaling shared data improves outdoor navigation, driving, and aerial instruction following, while routine indoor performance saturates earlier.
The driving safety gains are consistent with transfer of traversal experience, and higher aerial task completion aligns with supervision for stopping near the goal.
Small motions near the end of point-goal demonstrations may also shorten predicted driving paths, illustrating how the value of shared experience depends on the target task's execution requirements.
Corrective demonstrations complement this transfer by supervising recovery at policy-visited states, improving ground navigation with the same policy backbone.
When target data are scarce, accessible auxiliary experience can extend coverage of the behaviors the target task needs.
Task-specific scaling curves help identify where more routine demonstrations still improve performance. As these gains diminish, additional data collection can focus on recurring failures and the actions needed to recover from them.
The resulting policies achieve strong results across ABotN-PointBench, NAVSIM, and OpenFly, with demonstrations on quadrupeds, UAVs, and real driving recordings.
Cross-embodiment scaling is shaped by both the amount of experience and its relevance to the behaviors and states encountered during execution.

\RemoveFromHook{env/tabular*/begin}[navibridge]
\label{page:main-text-end}

\clearpage
\newpage

\section*{Acknowledgments}
We thank the WisdomStride Team at the School of Artificial Intelligence, Xi’an Jiaotong University, for providing robot platforms, technical support, and access to testing facilities for our real-world experiments.

\bibliographystyle{unsrt}
\bibliography{main}

@article{sun2026ego4wam,
  title={Ego4WAM: What Matters When Scaling Egocentric Human Data for Robot Learning?},
  author={Sun, Zhihao and Liu, Liu and Wang, Xinjiang and Jiang, Haoyi and Feng, Wei and Zhang, Huiqiang and Jia, Xiaosong and Su, Zhizhong and Wu, Zuxuan},
  journal={arXiv preprint arXiv:2609.40341},
  year={2026}
}

@article{suomela2026navigation,
  title={Data scaling for navigation in unknown environments},
  author={Suomela, Lauri and Takahata, Naoki and Arachchige, Sasanka Kuruppu and Edelman, Harry and K{\"a}m{\"a}r{\"a}inen, Joni-Kristian},
  journal={IEEE Robotics and Automation Letters},
  volume={11},
  number={5},
  pages={6114--6121},
  year={2026},
  publisher={IEEE}
}

@article{yang2024cross,
  title={Pushing the limits of cross-embodiment learning for manipulation and navigation},
  author={Yang, Jonathan and Glossop, Catherine and Bhorkar, Arjun and Shah, Dhruv and Vuong, Quan and Finn, Chelsea and Sadigh, Dorsa and Levine, Sergey},
  journal={arXiv preprint arXiv:2402.19432},
  year={2024}
}

@inproceedings{shah2023gnm,
  title={Gnm: A general navigation model to drive any robot},
  author={Shah, Dhruv and Sridhar, Ajay and Bhorkar, Arjun and Hirose, Noriaki and Levine, Sergey},
  booktitle={2023 IEEE International Conference on Robotics and Automation (ICRA)},
  pages={7226--7233},
  year={2023},
  organization={IEEE}
}

@inproceedings{ross2011dagger,
  title={A reduction of imitation learning and structured prediction to no-regret online learning},
  author={Ross, St{\'e}phane and Gordon, Geoffrey and Bagnell, Drew},
  booktitle={Proceedings of the fourteenth international conference on artificial intelligence and statistics},
  pages={627--635},
  year={2011},
  organization={JMLR Workshop and Conference Proceedings}
}

@article{dauner2024navsim,
  title={Navsim: Data-driven non-reactive autonomous vehicle simulation and benchmarking},
  author={Dauner, Daniel and Hallgarten, Marcel and Li, Tianyu and Weng, Xinshuo and Huang, Zhiyu and Yang, Zetong and Li, Hongyang and Gilitschenski, Igor and Ivanovic, Boris and Pavone, Marco and others},
  journal={Advances in Neural Information Processing Systems},
  volume={37},
  pages={28706--28719},
  year={2024}
}

@article{cao2025pseudo,
  title={Pseudo-simulation for autonomous driving},
  author={Cao, Wei and Hallgarten, Marcel and Li, Tianyu and Dauner, Daniel and Gu, Xunjiang and Wang, Caojun and Miron, Yakov and Aiello, Marco and Li, Hongyang and Gilitschenski, Igor and others},
  journal={arXiv preprint arXiv:2506.04218},
  year={2025}
}

@misc{navsim2026metrics,
  title = {The Extended Predictive Driver Model Score ({EPDMS})},
  author = {{NAVSIM Contributors}},
  howpublished = {Official NAVSIM repository documentation},
  url = {https://github.com/autonomousvision/navsim/blob/main/docs/metrics.md},
  year = {n.d.},
  note = {Accessed November 4, 2026}
}

@misc{nvidia2025cosmospredict2,
  title = {{Cosmos-Predict2}: World Simulation Model for Physical {AI}},
  author = {{NVIDIA}},
  year = {2025},
  url = {https://github.com/nvidia-cosmos/cosmos-predict2},
}

@article{kim2026cosmospolicy,
  title={Cosmos policy: Fine-tuning video models for visuomotor control and planning},
  author={Kim, Moo Jin and Gao, Yihuai and Lin, Tsung-Yi and Lin, Yen-Chen and Ge, Yunhao and Lam, Grace and Liang, Percy and Song, Shuran and Liu, Ming-Yu and Finn, Chelsea and others},
  journal={arXiv preprint arXiv:2601.16163},
  year={2026}
}

@inproceedings{gao2026openfly,
  title={Openfly: A comprehensive platform for aerial vision-language navigation},
  author={Gao, Yunpeng and Li, Chenhui and You, Zhongrui and Liu, Junli and Zhen, Li and Chen, Pengan and Chen, Qizhi and Tang, Zhonghan and Wang, Liansheng and Tang, Yiwen and others},
  booktitle={International Conference on Learning Representations},
  volume={2026},
  pages={101858--101871},
  year={2026}
}

@inproceedings{miao2026towards,
  title={Towards physically executable 3d gaussian for embodied navigation},
  author={Miao, Bingchen and Wei, Rong and Ge, Zhiqi and Gao, Shiqi and Zhu, Jingzhe and Wang, Renhan and Tang, Siliang and Xiao, Jun and Tang, Rui and Li, Juncheng and others},
  booktitle={International Conference on Learning Representations},
  volume={2026},
  pages={22945--22965},
  year={2026}
}

@article{gong2026abotn1,
  title={Abot-n1: Toward a general visual language navigation foundation model},
  author={Gong, Ruiyan and Guo, Yingnan and Hu, Junjun and Kong, Jintao and Leng, Xiaoxu and Li, Tianlun and Li, Weize and Liu, Fei and Liu, Zhicheng and Lu, Jia and others},
  journal={arXiv preprint arXiv:2607.10383},
  year={2026}
}

@article{xiong2026terratransfer,
  title={TerraTransfer++: Demonstration-Free End-to-End Planner},
  author={Xiong, Zikang and Li, Weixin and Wu, Zhouchonghao and Rangesh, Akshay and Bonde, Saarth and Hall, Grantland and Tang, Chen and Hu, Yihan and Zhan, Wei},
  journal={arXiv preprint arXiv:2610.30319},
  year={2026}
}

@article{chen2026fpddrive,
  title={FPD-Drive: Future Policy Distillation for End-to-End Autonomous Driving},
  author={Chen, Li-Heng and Chen, Qifeng and Wu, Chengwei and Zhang, Yingjie and Yang, Sheng and Fu, Hongbo and Ren, Shaoqing},
  journal={arXiv preprint arXiv:2610.13232},
  year={2026}
}

@article{xu2026toad,
  title={Test-Time Trajectory Optimization for Autonomous Driving},
  author={Xu, Yihong and Zablocki, Eloi and Yin, Yuan and Ramzi, Elias and Kirby, Ellington and Boulch, Alexandre and Cord, Matthieu},
  journal={arXiv preprint arXiv:2606.07170},
  year={2026}
}

@article{hong2026drivefuture,
  title={DriveFuture: Future-Aware Latent World Models for Autonomous Driving},
  author={Hong, Yufeng and Zhou, Xiaotian and Li, Yingyan and Zhou, Xiangpo and Liu, Lin and Luo, Yadan and Xu, Shaoqing and Yang, Lei and Song, Ziying},
  journal={arXiv preprint arXiv:2605.09701},
  year={2026}
}

@article{yuan2026suv,
  title={Suv: Future scene understanding as video generation for end-to-end driving},
  author={Yuan, Yibo and Fu, Jiacheng and Zhu, Jiangtong and Li, Yi and Han, Jianhua and Tian, Meng and Liu, Zhuohan and Xiong, Zhiwei and Xu, Hang and Fang, Jianwu and others},
  journal={arXiv preprint arXiv:2608.03084},
  year={2026}
}

@article{zhao2026simwam,
  title={Simwam: A simple world action model for end-to-end autonomous driving},
  author={Zhao, Zongchuang and Zhou, Xin and Xu, Tianyang and Sun, Zhengyang and Zhou, Kaixuan and Wu, Yu and Li, Honglin and Liang, Dingkang and Bai, Xiang},
  journal={arXiv preprint arXiv:2608.07468},
  year={2026}
}

@article{liang2026vital,
  title={VITAL: From Visual Motion to Precise Action},
  author={Liang, Ao and Liu, Youquan and Li, Mingxiao and Chow, Wei and Kong, Lingdong and Liao, Yue and Lu, Dongyue and Yang, Yu and Yin, Wei and Chen, Runnan},
  journal={arXiv preprint arXiv:2610.20604},
  year={2026}
}

@article{yang2026sparsewam,
  title={SparseWAM: Driving World-Action Models are Worth 16 Register Tokens},
  author={Yang, Yixiang and Zhang, Lizhong and Zhang, Xingyu and Chen, Zhiwen and Ling, Jie and Yu, Qiankun and Liu, Xingliang and Lei, Yang and Zhang, Zhipeng},
  journal={arXiv preprint arXiv:2611.00044},
  year={2026}
}

@article{zhu2026fsdvln,
  title={FSD-VLN: Fast-Slow Dual-System Modeling for Aerial Long-Horizon Vision-Language Navigation},
  author={Zhu, Xueke and Meng, Qingyan and Yu, Liutao and Zhang, Wei and Ma, Zhengyu and Zhou, Huihui and Tian, Yonghong},
  journal={arXiv preprint arXiv:2607.08359},
  year={2026}
}

@article{chu2026abotn0,
  title={Abot-n0: Technical report on the vla foundation model for versatile embodied navigation},
  author={Chu, Zedong and Xie, Shichao and Wu, Xiaolong and Shen, Yanfen and Luo, Minghua and Wang, Zhengbo and Liu, Fei and Leng, Xiaoxu and Hu, Junjun and Yin, Mingyang and others},
  journal={arXiv preprint arXiv:2602.11598},
  year={2026}
}

@article{xie2026sparse,
  title={Sparse-WAM: Accelerating World Action Models via Action-Guided Sparse Imagination},
  author={Xie, Xinling and Wang, Haodong and Mi, Jiazhi and Liu, Zhiming and Hong, Zicong and Pang, Xiaoyi and Liu, Qianli and Hu, Yangjia and Chen, Ying and Yan, Zhengyang and others},
  journal={arXiv preprint arXiv:2609.38984},
  year={2026}
}

@inproceedings{cheng2025navila,
  title = {{NaVILA}: Legged Robot Vision-Language-Action Model for Navigation},
  author = {Cheng, An-Chieh and Ji, Yandong and Yang, Zhaojing and Gongye, Zaitian and Zou, Xueyan and Kautz, Jan and Biyik, Erdem and Yin, Hongxu and Liu, Sifei and Wang, Xiaolong},
  booktitle = {Robotics: Science and Systems},
  year = {2025},
  url = {https://www.roboticsproceedings.org/rss21/p018.html}
}

@inproceedings{zhang2024navid,
  title = {{NaVid}: Video-based {VLM} Plans the Next Step for Vision-and-Language Navigation},
  author = {Zhang, Jiazhao and Wang, Kunyu and Xu, Rongtao and Zhou, Gengze and Hong, Yicong and Fang, Xiaomeng and Wu, Qi and Zhang, Zhizheng and Wang, He},
  booktitle = {Robotics: Science and Systems},
  year = {2024},
  url = {https://www.roboticsproceedings.org/rss20/p079.html}
}

@inproceedings{savva2019habitat,
  title={Habitat: A platform for embodied ai research},
  author={Savva, Manolis and Kadian, Abhishek and Maksymets, Oleksandr and Zhao, Yili and Wijmans, Erik and Jain, Bhavana and Straub, Julian and Liu, Jia and Koltun, Vladlen and Malik, Jitendra and others},
  booktitle={2019 IEEE/CVF International Conference on Computer Vision (ICCV)},
  pages={9338--9346},
  year={2019},
  organization={IEEE}
}

@article{wijmans2020ddppo,
  title={Dd-ppo: Learning near-perfect pointgoal navigators from 2.5 billion frames},
  author={Wijmans, Erik and Kadian, Abhishek and Morcos, Ari and Lee, Stefan and Essa, Irfan and Parikh, Devi and Savva, Manolis and Batra, Dhruv},
  journal={arXiv preprint arXiv:1911.00357},
  year={2019}
}

@inproceedings{shah2023vint,
  title = {{ViNT}: A Foundation Model for Visual Navigation},
  author = {Shah, Dhruv and Sridhar, Ajay and Dashora, Nitish and Stachowicz, Kyle and Black, Kevin and Hirose, Noriaki and Levine, Sergey},
  booktitle = {Conference on Robot Learning},
  pages = {711--733},
  year = {2023},
  url = {https://proceedings.mlr.press/v229/shah23a.html}
}

@inproceedings{openx2024,
  title={Open x-embodiment: Robotic learning datasets and rt-x models: Open x-embodiment collaboration 0},
  author={O’Neill, Abby and Rehman, Abdul and Maddukuri, Abhiram and Gupta, Abhishek and Padalkar, Abhishek and Lee, Abraham and Pooley, Acorn and Gupta, Agrim and Mandlekar, Ajay and Jain, Ajinkya and others},
  booktitle={2024 IEEE International Conference on Robotics and Automation (ICRA)},
  pages={6892--6903},
  year={2024},
  organization={IEEE}
}

@inproceedings{lin2025datascaling,
  title={Data scaling laws in imitation learning for robotic manipulation},
  author={Lin, Fanqi and Hu, Yingdong and Sheng, Pingyue and Wen, Chuan and You, Jiacheng and Gao, Yang},
  booktitle={International Conference on Learning Representations},
  volume={2025},
  pages={54877--54910},
  year={2025}
}

@inproceedings{laskey2017dart,
  title={Dart: Noise injection for robust imitation learning},
  author={Laskey, Michael and Lee, Jonathan and Fox, Roy and Dragan, Anca and Goldberg, Ken},
  booktitle={Conference on robot learning},
  pages={143--156},
  year={2017},
  organization={PMLR}
}

@inproceedings{bansal2019chauffeurnet,
  title = {Chauffeurnet: Learning to drive by imitating the best and synthesizing the worst},
  author = {Bansal, Mayank and Krizhevsky, Alex and Ogale, Abhijit},
  booktitle = {Robotics: Science and Systems},
  year = {2019},
  url = {https://www.roboticsproceedings.org/rss15/p31.html}
}

@inproceedings{liu2023aerialvln,
  title={Aerialvln: Vision-and-language navigation for uavs},
  author={Liu, Shubo and Zhang, Hongsheng and Qi, Yuankai and Wang, Peng and Zhang, Yanning and Wu, Qi},
  booktitle={2023 IEEE/CVF International Conference on Computer Vision (ICCV)},
  pages={15338--15348},
  year={2023},
  organization={IEEE}
}

@inproceedings{zhu2025uwm,
  title = {Unified World Models: Coupling Video and Action Diffusion for Pretraining on Large Robotic Datasets},
  author = {Zhu, Chuning and Yu, Raymond and Feng, Siyuan and Burchfiel, Benjamin and Shah, Paarth and Gupta, Abhishek},
  booktitle = {Robotics: Science and Systems},
  year = {2025},
  url = {https://www.roboticsproceedings.org/rss21/p015.html}
}

\clearpage
% Keep report page numbering continuous across the source appendix's reset.
\edef\firstsupplementpage{\number\value{page}}
% Fit natural-width supplementary tables within the report margins.
\AddToHook{env/tabular/begin}[navibridge]{\setlength{\tabcolsep}{.55\tabcolsep}}
\clearpage
\appendix
\setcounter{page}{1}
\setcounter{figure}{0}
\setcounter{table}{0}
\renewcommand{\thefigure}{S\arabic{figure}}
\renewcommand{\thetable}{S\arabic{table}}
\renewcommand{\theHfigure}{supp.\arabic{figure}}
\renewcommand{\theHtable}{supp.\arabic{table}}
\maketitlesupplementary

\section{Training Setup}
\label{app:training}

All tasks share a world-action model architecture with a Cosmos-Predict2-2B-Video2World backbone.
Point-goal navigation uses a numerical goal, aerial navigation uses a language instruction, and driving uses a navigation command.
\Cref{tab:training-recipes} lists the recipes for source composition, policy correction, and the comparison at matched driving draws.

\begin{table*}[t]
    \caption{Initialization, source proportions, and peak learning rates for single-task, joint, and corrective training. Single-source sampling weights are normalized to one.}
    \label{tab:training-recipes}
    \centering
    \small
    \begin{tabularx}{\textwidth}{@{}p{0.20\textwidth}p{0.21\textwidth}Xr@{}}
        \toprule
        Recipe & Initialization & Data sources and sampling proportions & Learning rate \\
        \midrule
        Indoor only & Pretrained & Indoor 1.0 & 0.0001 \\
        Outdoor only & Pretrained & Outdoor 1.0 & 0.0001 \\
        UAV only & Pretrained & OpenFly 1.0 & 0.0001 \\
        Driving only & Pretrained & NAVSIM 1.0 & 0.0001 \\
        Joint ground & Pretrained & Outdoor 0.6, indoor 0.4 & 0.0001 \\
        NaviBridge (four sources) & Pretrained & Outdoor 0.40, indoor 0.35, UAV 0.20, driving 0.05 & 0.0001 \\
        Indoor DAgger & Joint ground, 12k & Indoor base 0.8, indoor DAgger 0.2 & 0.00002 \\
        Outdoor DAgger & Joint ground, 12k & Outdoor base 0.8, outdoor DAgger 0.2 & 0.00002 \\
        Joint DAgger & Joint ground, 12k & Indoor base 0.1, indoor DAgger 0.3, outdoor base 0.3, outdoor DAgger 0.3 & 0.00001 \\
        Historical joint DAgger & Joint ground, 12k & Indoor base 0.1, indoor DAgger 0.6, outdoor base 0.1, outdoor DAgger 0.2 & 0.00001 \\
        \bottomrule
    \end{tabularx}
\end{table*}

\paragraph{Standard training settings.}
Indoor, outdoor, UAV, and driving single-task training, joint ground training, and four-source joint training use the same AdamW optimizer and learning-rate schedule.
The peak learning rate is 0.0001, with weight decay 0.1, momentum coefficients 0.9 and 0.99, and epsilon $10^{-8}$.
The learning rate warms up linearly for 1k steps and decays linearly to 0.3 times its peak within the first 20k-step cycle, after which it remains at 0.06 times the peak.
All stages use one gradient accumulation step.

All standard models initialize from pretrained Cosmos-Predict2-2B weights.
The UAV, indoor, outdoor, driving, and both standard joint recipes use a batch size of 36 per device.

Outdoor and indoor DAgger provide about 199.7k and 83.6k windows, respectively.
The three correction recipes in \cref{tab:correction} initialize from the joint ground checkpoint at 12k steps and restart optimization.
\Cref{tab:correction} uses one set of weights per recipe to evaluate both environments.

\paragraph{Action normalization.}
The source-specific spatial normalization coefficients $\kappa_s$ in \cref{eq:actions} are $0.30\,\mathrm{m}$ for indoor navigation, $0.60\,\mathrm{m}$ for outdoor navigation, approximately $2.696\,\mathrm{m}$ for UAV instruction following, and $3.50\,\mathrm{m}$ for driving.

\paragraph{Driving and total sample draws.}
Driving-only training and NaviBridge use the same corpus of about 53.1k driving samples and a batch size of 36 per device.
They use 32 and 96 data-parallel devices, respectively, for effective global batch sizes of 1,152 and 3,456.
The driving sampling proportions are 1.0 and 0.05, respectively.
The cumulative expected driving draws in \cref{fig:drive-draws} are
\begin{equation}
    \widehat{N}_{\mathrm{drive}}(s)=s\,B\,p_{\mathrm{drive}},
    \label{eq:driving-draws}
\end{equation}
where $s$ is the number of optimizer updates, $B$ is the effective global batch size, and $p_{\mathrm{drive}}$ is the driving sampling proportion.
Repeated uses of a sample count as separate draws. The driving-only and joint recipes draw 1,152 and an expected 172.8 driving samples per update, respectively.
At the comparison point, the driving-only run uses 1.728M draws in total. The joint run uses 34.56M draws across all sources, including 1.728M expected driving draws.
The best evaluated checkpoints use about 3.456M and 1.555M driving draws, respectively. The joint checkpoint therefore uses about 45\% as many driving draws.

\paragraph{Corpus sizes.}
Corpus sizes count distinct training samples, while cumulative draws count their repeated use.
The base ground corpus contains about 6.69M samples.
Across base training and correction, the indoor, outdoor, and joint recipes use about 6.78M, 6.89M, and 6.98M samples, respectively.
The full four-source NaviBridge corpus contains about 8.61M samples.
ABot-N1 pretraining uses approximately 30M samples across its navigation tasks, with about 13.3M for the slow system and 16.4M for the fast system~\cite{gong2026abotn1}.

\section{Evaluation Protocols}
\label{app:protocols}

\subsection{ABotN-PointBench}
\label{app:ground-protocol}

The ground navigation evaluations and NaviBridge benchmark runs use evaluation seed 42 and two denoising steps.
ABotN-PointBench includes 240 fixed indoor tasks, with 120 easy and 120 hard tasks.
The 225 outdoor tasks comprise 75 tasks in each of the short, medium, and long distance groups.
Success requires a final distance of at most 0.5\,m and fewer than one collision indoors or fewer than three collisions outdoors.
SPL weights path efficiency by the same success criterion, and paired trajectory comparisons match task identity, initial distance, and shortest-path length.

Correction trajectories are collected independently in training scenes.
The indoor and outdoor correction samples cover 418 and 25 scenes, respectively, with no overlap with the 16 indoor and 15 outdoor test scenes.
The initial positions, goals, and recovery states of correction tasks all come from these training scenes, while evaluation uses fixed tasks in the test scenes.

\subsection{OpenFly}
\label{app:openfly-protocol}

The policy outputs horizontal waypoints, and the simulator uses flight altitudes from the reference trajectory in the main experiments.
An episode succeeds if its actual endpoint is within 20\,m of the goal in three dimensions, with a strict distance threshold.
NaviBridge signals a stop when its predicted net horizontal displacement is less than 1\,m.

To examine whether joint training gains depend on reference altitude changes, we count tasks with such changes as failures and retain them in the denominator.
The remaining trajectories maintain their initial altitude throughout and use the same termination conditions for scoring.
This procedure gives the additional column in \cref{tab:full-scaling} and the comparisons in \cref{app:openfly-public,app:openfly-joint}.

\subsection{NAVSIM}
\label{app:navsim-protocol}

We use the NAVSIM V2 navhard split and two-stage EPDMS with reactive traffic agents.
The model receives the current front-camera image, four images from the same camera's history, and a navigation command.
The five output waypoints correspond to 0.5, 1, 2, 3, and 4 seconds into the future.
For scoring, we linearly interpolate them to eight positions spaced 0.5 seconds apart.

\section{Data Scaling Experiments}
\label{app:scaling}

\subsection{Stratified Trajectory Sampling}
\label{app:subset-sampling}

The four data scales use the same sources and sampling proportions, with nested subsets containing 25\%, 50\%, 75\%, and 100\% of each source.
Indoor, outdoor, and OpenFly data are sampled by complete trajectory, and NAVSIM is sampled by complete driving log.
All training windows from the same trajectory or log are retained or excluded together at each scale.

We stratify outdoor data by category and scene, indoor data by scene, OpenFly by environment and trajectory category, and NAVSIM by city.
Within each stratum, we first sort task identifiers and then shuffle them once using a random seed.
Following this order, we accumulate the number of windows per task and retain a prefix at each target fraction.
For a task whose window interval crosses the fraction cutoff, we determine inclusion using the midpoint of that cumulative interval.
This procedure approximately preserves the target fraction of training windows in each stratum; differences in complete trajectory lengths cause small deviations in the realized fractions.

The three subsets share the same random order and therefore satisfy
\begin{equation}
    \mathcal{D}_{25\%}\subset\mathcal{D}_{50\%}
    \subset\mathcal{D}_{75\%}\subset\mathcal{D}_{100\%}.
\end{equation}

The upper part of \cref{tab:full-scaling} lists the strata, task counts, and windows at each scale; the lower part reports evaluation results.
The realized 25\% window fractions for the four sources are 24.91\%, 25.04\%, 25.00\%, and 25.11\%, respectively.
Here k and M denote thousands and millions, and the table rounds the counts.
The 25\% indoor subset covers 785 scenes, while the other indoor scales cover all 790 scenes.

\subsection{Results at Each Data Scale}

\paragraph{Checkpoint selection.}
All four data-scale variants use the same checkpoint selection criterion.
We independently hold out a validation set from the training corpus, with no overlap with test trajectories.
For each scale, we compute the mean squared error (MSE) averaged over the validation set for every candidate checkpoint and select the checkpoint with the lowest average MSE for final testing.

\begin{table*}[t]
    \caption{Nested training subset sizes and task-level results. The upper panel gives sampling units and training window counts; the lower panel gives evaluation results at each scale. Ground means average indoor and outdoor SR$_{\mathrm{new}}$ and SPL$_{\mathrm{new}}$ separately. Driving uses the two-stage combined NAVSIM EPDMS. All four scales use the same 1,795 OpenFly tasks from eight scenes, with equal task weights under both scoring rules. \Cref{fig:scaling} uses the reference-altitude-assisted column.}
    \label{tab:full-scaling}
    \centering
    \small
    \setlength{\tabcolsep}{3pt}
    \begin{tabularx}{\textwidth}{@{}lXXrrrr@{}}
        \toprule
        Source & Strata (full count) & Sampling units (full count) & 25\% & 50\% & 75\% & 100\% \\
        \midrule
        Outdoor PointGoal & Category / scene (66) & Trajectories (80.014k) & 1.518M & 3.043M & 4.563M & 6.094M \\
        Indoor PointGoal & Scenes (790) & Trajectories (13.530k) & 150.058k & 300.268k & 449.296k & 599.360k \\
        OpenFly & Environment / category (97) & Trajectories (97.400k) & 465.692k & 931.530k & 1.397M & 1.863M \\
        NAVSIM & Cities (4) & Logs (1.137k) & 13.331k & 26.599k & 39.888k & 53.092k \\
        \bottomrule
    \end{tabularx}
    \par\medskip
    \begin{tabular}{@{}lrrrrcc@{}}
        \toprule
        & \multicolumn{3}{c}{ABotN-PointBench SR$_{\mathrm{new}}$ / SPL$_{\mathrm{new}}$ (\%)}
        & NAVSIM & \multicolumn{2}{c}{OpenFly SR (\%)} \\
        \cmidrule(lr){2-4}\cmidrule(lr){6-7}
        Data & Indoor & Outdoor & Mean & EPDMS & Reference altitude & Altitude changes fail \\
        \midrule
        25\% & 88.75 / 85.48 & 87.11 / 85.85 & 87.93 / 85.66 & 32.42 & 41.56 & 39.72 \\
        50\% & 94.17 / 90.51 & 88.89 / 88.35 & 91.53 / 89.43 & 36.55 & 43.79 & 38.16 \\
        75\% & 95.83 / 91.77 & 90.22 / 88.63 & 93.03 / 90.20 & 37.12 & 52.03 & 46.52 \\
        100\% & 95.42 / 91.90 & 93.33 / 92.32 & 94.38 / 92.11 & 40.74 & 55.43 & 50.58 \\
        \bottomrule
    \end{tabular}
\end{table*}

The dashed curves in \cref{fig:scaling} use a shared empirical power-law form.
For each series, we estimate the coefficient and exponent by least-squares regression on the logarithms of the data fraction and the gap to the maximum score.
Both SR and EPDMS have a maximum of 100, and each fit uses the four measured values.

\section{Task-Level Ground Navigation Analysis}
\label{app:ground-analysis}

\subsection{Indoor Navigation}

\begin{figure*}[t]
    \centering
    \includegraphics[width=\textwidth]{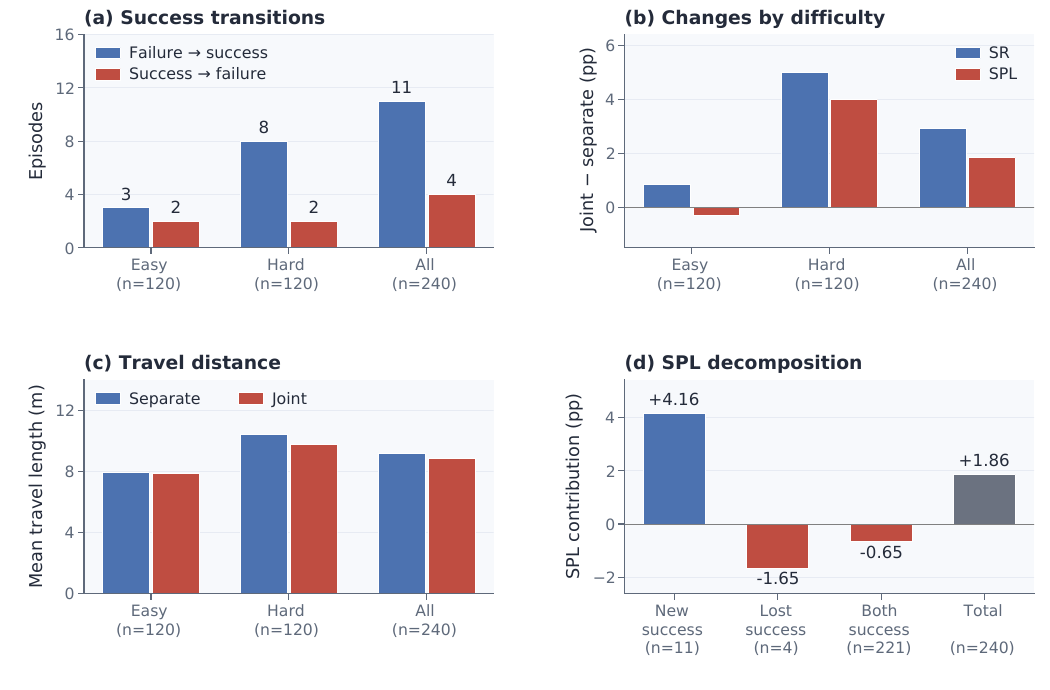}
    \caption{Task-level analysis of joint indoor and outdoor training on the indoor benchmark. Each difficulty group contains 120 tasks. The SPL decomposition normalizes every term by the full set of 240 tasks.}
    \label{fig:indoor-analysis}
\end{figure*}

The indoor-only and joint models both succeed on 221 tasks and both fail on four.
Joint training converts 11 failures to successes and four successes to failures.
Both models reach the goal on every task, so the collision criterion determines all 15 changes in success.
Collision counts fall to zero on the recovered tasks and rise from zero to between three and six on the lost tasks.
The main benefit of cross-environment training is safer navigation completion.

On hard tasks, successes increase from 108 to 114, accounting for six of the seven net additional successes.
On easy tasks they increase from 117 to 118.
Hard-task SPL rises from 86.21\% to 90.22\%, while easy-task SPL changes from 94.52\% to 94.22\%, concentrating the gains on harder tasks.

Let $G$ be tasks only the joint model completes successfully, $F$ those only the single-domain model completes successfully, and $B$ those both complete successfully.
With $n=240$ and per-task path efficiency $p_i$, we decompose the overall SPL change as
\begin{equation}
    \begin{aligned}
    \Delta\mathrm{SPL}={}&\frac{1}{n}\sum_{i\in G}p_i^{\mathrm{joint}}
        -\frac{1}{n}\sum_{i\in F}p_i^{\mathrm{single}}\\
        &+\frac{1}{n}\sum_{i\in B}
        \left(p_i^{\mathrm{joint}}-p_i^{\mathrm{single}}\right).
    \end{aligned}
    \label{eq:spl-decomposition}
\end{equation}
Per-task SPL combines success with path efficiency, so tasks that fail under both models contribute zero.
The recovered, lost, and common-success terms contribute $+4.1557$, $-1.6484$, and $-0.6465$ percentage points, respectively, for a total of $+1.8608$.
Recovered successes account for the overall improvement, while mean path efficiency drops on tasks both models complete.

\subsection{Outdoor Navigation}

Across 225 outdoor tasks, the outdoor-only and joint models succeed on 201 and 211, respectively.
Joint training converts 14 failures to successes and four successes to failures.
Recovered, lost, and common-success tasks contribute $+6.1855$, $-1.7567$, and $-0.361474$ percentage points to SPL, for a total of $+4.0674$.

\section{NAVSIM Driving Metrics}
\label{app:driving-analysis}

\begin{figure*}[t]
    \centering
    \includegraphics[width=\textwidth]{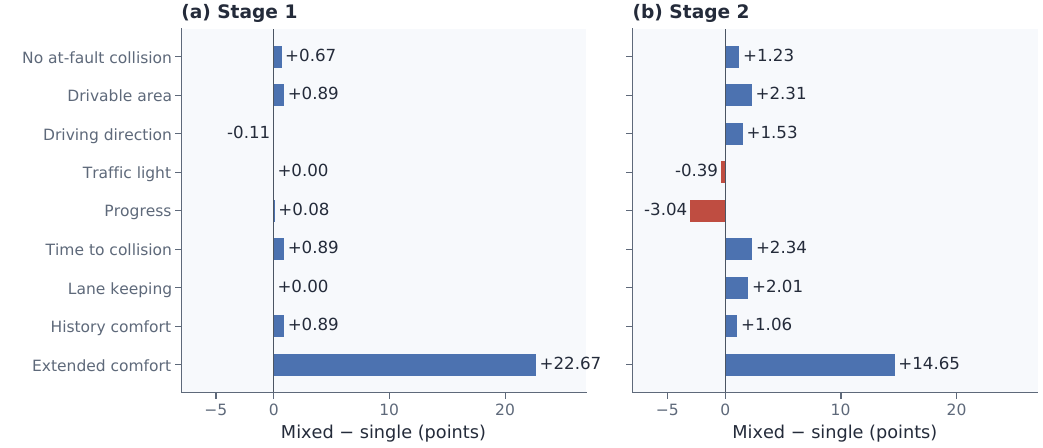}
    \caption{NAVSIM navhard component scores for the driving-only model and NaviBridge at their respective best checkpoints. Stage 2 weights subsequent observations using the NAVSIM protocol.}
    \label{fig:driving-components}
\end{figure*}

\Cref{fig:driving-components} compares driving safety, road compliance, progress, and comfort.
Relative to driving-only training, NaviBridge keeps Stage 1 progress nearly unchanged ($+0.08$ points) and improves Extended Comfort by 22.67 points.
In Stage 2, the scores for no at-fault collisions, drivable-area compliance, and TTC improve by 1.23, 2.31, and 2.34 points, respectively.
Extended Comfort improves by 14.65 points, and road progress decreases by 3.04 points.

The four-constraint pass rate in \cref{sec:driving-behavior} gives equal weight to the original predictions on 5,462 paired Stage 2 observations.
A trajectory passes when no at-fault collisions (NC), drivable-area compliance (DAC), driving-direction compliance (DDC), and traffic-light compliance (TLC) all equal one.
We calculate the 95\% confidence interval for the difference using 10,000 bootstrap resamples of 225 paired scenario groups.
All subsequent observations in a sampled group are included together.

With the same equal weighting, the Stage 2 progress difference is $-1.29$ points.
On the 2,988 Stage 2 observations where both models pass all four constraints, NaviBridge's four-second predicted paths are 0.22\,m (1.7\%) shorter on average than those of the driving-only model.
The Spearman correlation between progress differences and predicted path length differences on these observations is 0.904.

In Stage 2, NaviBridge has higher drivable-area compliance, shorter predicted paths, and lower road progress.
A shorter trajectory may end before reaching a section where the other vehicle leaves the road, raising its compliance score.
We therefore compare the two trajectories at the same road locations to examine NaviBridge's behavior as it actually passes these sections.
We define a road departure as any corner of the vehicle bounding box lying outside the mapped drivable area during simulation, which fails the drivable-area compliance (DAC) criterion.
We select all paired observations in which only one vehicle departs, replay both original four-second predictions, and check whether the vehicle that stays within the drivable area drives past the other's first departure on the same route.

To locate the departure, we project the vehicle center onto the route centerline in the first frame with a corner outside the boundary and in the preceding frame (0.1\,s earlier).
We count the other vehicle as \emph{past the site} if its center travels at least 0.5\,m beyond both positions on the same road segment or junction connector. It \emph{falls short} if it remains at least 0.5\,m before both. We leave cases within the margin or on ambiguous branches unresolved.

\begin{table}[htbp]
    \caption{Paired road-departure analysis on NAVSIM navhard. Counts show whether the on-road model passes the other model's first departure site along the same route.}
    \label{tab:departure-sites}
    \centering
    \small
    \setlength{\tabcolsep}{2pt}
    \begin{tabular*}{\columnwidth}{@{\extracolsep{\fill}}clrrrr@{}}
        \toprule
        Stage & On-road model & Total & Past site & Fell short & Unresolved \\
        \midrule
        \multirow{2}{*}{2} & NaviBridge    & 312 & 203 & 25 & 84 \\
          & Driving-only & 204 & 125 & 12 & 67 \\
        \addlinespace
        \multirow{2}{*}{1} & NaviBridge    & 35  & 25  & 1  & 9  \\
          & Driving-only & 31  & 28  & 1  & 2  \\
        \bottomrule
    \end{tabular*}
\end{table}

In Stage 2, NaviBridge stayed within the drivable area in 312 observations where the driving-only vehicle crossed the boundary.
In 203 of these observations (65.1\%), NaviBridge drove past the driving-only vehicle's first departure along the same road, keeping all four corners inside the drivable area.
In the reverse comparison, the driving-only vehicle passed NaviBridge's first departure in 125 of the 204 observations where only the driving-only vehicle stayed on the road.

On road sections reached by both vehicles, NaviBridge drove past the departure site while staying within the drivable area in 78 more observations than the driving-only model (203 versus 125).
These cases account for most of the 108 net gains in Stage 2 drivable-area compliance.
The trajectories show NaviBridge following more compliant paths through the same road sections and keeping its vehicle within the drivable area more reliably.

\begin{figure}[t]
    \centering
    \includegraphics[width=0.8\columnwidth]{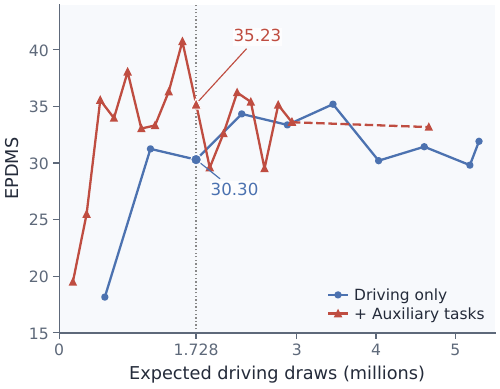}
    \caption{NAVSIM navhard EPDMS versus cumulative expected draws from the driving corpus. The dashed line marks 1.728M driving draws.}
    \label{fig:drive-draws}
\end{figure}

\paragraph{Performance at matched driving draws.}
At 1.728M expected draws from the same driving corpus, NaviBridge scores 35.23 EPDMS and the driving-only model scores 30.30 (\cref{fig:drive-draws}).
Across all sources, NaviBridge uses 34.56M draws at this comparison point (\cref{app:training}).

\section{Action Distributions and Spatial Behavior}
\label{app:actions}

\subsection{Action Statistics over the Full Corpus}

We analyze the full 8.61M-sample corpus spanning outdoor, indoor, UAV, and driving data.
Each sample contains five future action increments normalized by the source-specific scale in \cref{sec:representation}.
The normalized path length for window $x$ is
\begin{equation}
    \ell(x)=\sum_{k=1}^{5}\lVert a_k(x)\rVert_2.
    \label{eq:window-distance}
\end{equation}
We detect repeated waypoints by testing for a near-zero displacement between successive future waypoints,
\begin{equation}
    r(x)=\mathbf{1}\!\left[
        \min_{k\in\{2,3,4,5\}}\lVert a_k(x)\rVert_2 < 10^{-5}
    \right].
    \label{eq:repeat-window}
\end{equation}

Across the three auxiliary sources, every window meeting this criterion has a last action with zero displacement and repeated future images.
These windows reuse terminal observations and waypoints, whereas only 10 driving samples meet the zero-displacement criterion.
\Cref{tab:action-stats} summarizes these distributions with equal weight for every sample within each source.

\begin{table*}[t]
    \caption{Action distributions for the four complete training corpora, with sample counts rounded. Path length sums the magnitudes of the five normalized future increments. Repeated windows use the same zero-displacement criterion in all sources.}
    \label{tab:action-stats}
    \centering
    \small
    \begin{tabular}{@{}lrrrr@{}}
        \toprule
        Source & Training samples & Mean normalized length & Repeated-waypoint windows (\%) & Mean without repetitions \\
        \midrule
        Outdoor & 6.09M & 4.45 & 16.42 & 4.87 \\
        Indoor & 599k & 4.01 & 16.15 & 4.44 \\
        UAV & 1.86M & 6.01 & 20.78 & 6.88 \\
        Driving & 53.1k & 5.15 & 0.019 & 5.15 \\
        \bottomrule
    \end{tabular}
\end{table*}

Mean normalized paths are shorter than driving paths in indoor and outdoor data and longer in UAV data.
Mean displacement from the first to the fifth action falls from 0.98 to 0.82 outdoors, 0.89 to 0.73 indoors, and 1.38 to 1.03 in UAV data.
Driving has mean displacements of 0.74, 0.73, 1.34, 1.23, and 1.12 at its five action positions.
Driving labels cover several prediction horizons, with about 33.7k samples spanning approximately four seconds.
The increments reflect both motion and the time intervals between waypoints.

\begin{figure*}[t]
    \centering
    \includegraphics[width=\textwidth]{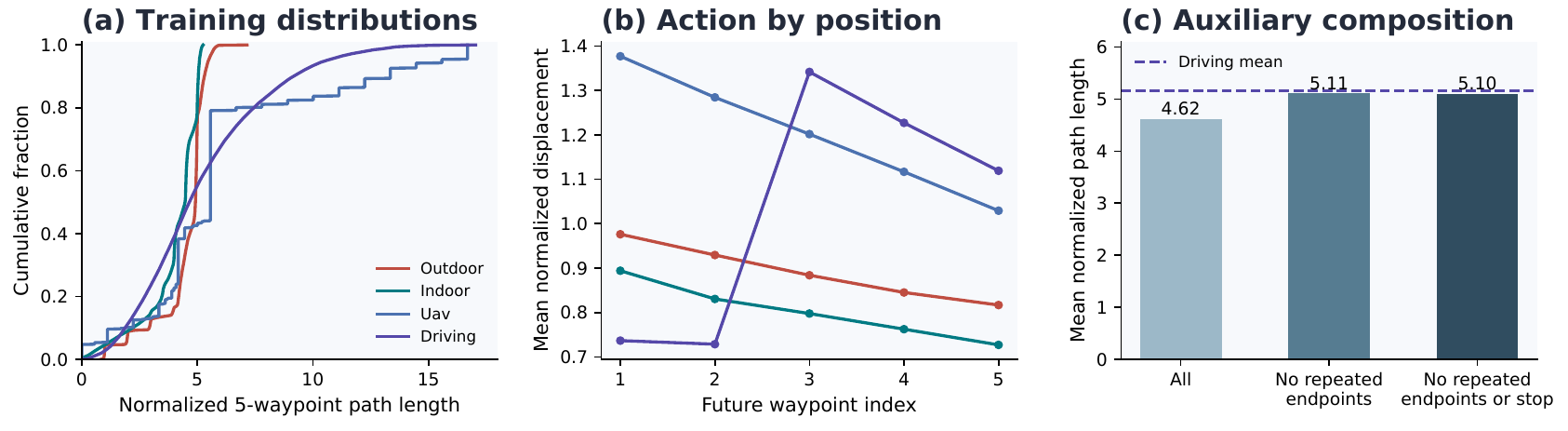}
    \caption{Action statistics and terminal supervision. (a) Cumulative distribution of normalized path length. (b) Mean displacement at each of the five future action positions. (c) Auxiliary mean path lengths for all windows and two subsets. Windows receive equal weight within each source. The combined statistics apply weights of 0.40, 0.35, and 0.20 to outdoor, indoor, and UAV data, respectively, then renormalize over retained windows. The first subset excludes repeated waypoints. The second also excludes stop or arrival labels. \Cref{app:terminal-windows} reports statistics under the training sampler.}
    \label{fig:action-distributions}
\end{figure*}

\subsection{Terminal Windows under the Training Sampler}
\label{app:terminal-windows}

We weight each window by its occurrence count under the joint training sampler.
With these weights, 17.64\% of auxiliary windows contain repeated terminal waypoints.
The mean normalized length is 4.690 for auxiliary data and 5.141 for driving.
Excluding repeated-waypoint windows changes the means to 5.207 and 5.142, respectively.
Indoor and outdoor point-goal data account for 72.91\% of windows with repeated terminal waypoints.
These examples supervise a gradual shortening of the predicted path near the goal.

\subsection{Obstacle Avoidance and Detour Experience}
\label{app:detours}

The outdoor corpus contains about 2.41M windows from the obstacle-avoidance stream.
They account for 39.48\% of the outdoor data and about 15.79\% of draws under the joint training mixture.
This stream supervises paths constrained by obstacles, complementing the connecting paths and turns elsewhere in the outdoor corpus.

The 599.360k indoor samples come from 13.530k source trajectories.
We link windows to trajectory geometry and measure the minimum obstacle distance and the ratio of path length to the straight-line distance between the full trajectory's endpoints.
Weighted by training window count, 93.37\% of windows come from trajectories passing within 0.5\,m of an obstacle.
The standard navigation category contains about 9.52k trajectories and 413k samples.
In this category, 54.02\% of windows come from trajectories more than 20\% longer than the straight-line distance.
With a 50\% detour threshold, the share of windows is 19.80\%.
These statistics characterize traversal through constrained spaces and detours in the auxiliary data.

Obstacle avoidance and detour supervision are consistent with the gains in collision safety and drivable-area compliance in the original driving predictions.
Repeating terminal waypoints changes the distribution of action magnitudes, consistent with shorter predicted driving paths and more frequent small displacements near aerial goals.
\Cref{app:openfly-behavior} gives the stopping criterion and paired trajectory results.

\section{OpenFly Evaluation}
\label{app:openfly}

\subsection{Benchmark Comparison}
\label{app:openfly-public}

\Cref{tab:openfly-public} summarizes results on the OpenFly Test Seen and Test Unseen splits.
The published OpenFly row weights the scores for corresponding publicly released scenes by their task counts.
For NaviBridge, tasks with changes in reference altitude contribute zero to SR, OSR, and SPL (\cref{app:openfly-protocol}).
Navigation error (NE) measures mean distance from the actual endpoint to the goal; we compute NaviBridge's NE over constant-altitude tasks.
Oracle success rate (OSR) measures the fraction of tasks that come within 20\,m of the goal in three dimensions at any point after the start.

\begin{table*}[t]
    \caption{Altitude-checked results and published baselines on the released OpenFly scenes. NaviBridge counts tasks involving reference altitude changes as failures. The OpenFly row aggregates published scores on the corresponding scenes; the other baselines retain their published values~\cite{gao2026openfly,zhu2026fsdvln}. Bold denotes the best SR, OSR, and SPL in each column.}
    \label{tab:openfly-public}
    \centering
    \small
    \setlength{\tabcolsep}{5pt}
    \begin{tabular}{@{}lrrrrrrrr@{}}
        \toprule
        & \multicolumn{4}{c}{Test Seen} & \multicolumn{4}{c}{Test Unseen} \\
        \cmidrule(lr){2-5}\cmidrule(lr){6-9}
        Method & NE (m) $\downarrow$ & SR (\%) $\uparrow$ & OSR (\%) $\uparrow$ & SPL (\%) $\uparrow$
               & NE (m) $\downarrow$ & SR (\%) $\uparrow$ & OSR (\%) $\uparrow$ & SPL (\%) $\uparrow$ \\
        \midrule
        Seq2Seq & 205 & 2.9 & 24.3 & 2.6 & 229 & 2.1 & 20.6 & 1.1 \\
        CMA & 161 & 5.4 & 28.1 & 4.8 & 217 & 4.6 & 24.4 & 2.1 \\
        See-Point-Fly & \textemdash & \textemdash & \textemdash & \textemdash & 191 & 8.2 & 12.7 & 6.3 \\
        AerialVLN & 139 & 7.5 & 30.0 & 6.8 & 214 & 7.3 & 28.1 & 4.4 \\
        Navid & 153 & 13.0 & 38.2 & 11.6 & 210 & 10.8 & 27.2 & 5.0 \\
        NaVila & 132 & 20.3 & 53.5 & 17.8 & 202 & 14.7 & 42.1 & 9.6 \\
        FSD-VLN & 76 & 26.7 & 43.3 & 22.8 & 78 & 13.6 & 28.4 & 10.7 \\
        OpenFly & 118.7 & 28.8 & 63.1 & 18.7 & 164.4 & 26.2 & \textbf{68.8} & 18.0 \\
        NaviBridge (ours) & 148.1 & \textbf{51.55} & \textbf{68.51} & \textbf{46.43} & 85.0 & \textbf{47.28} & 56.44 & \textbf{42.59} \\
        \bottomrule
    \end{tabular}
\end{table*}

Under this scoring rule, NaviBridge obtains the highest SR and SPL in the table on both splits.
The full-data joint model reaches 50.58\% overall SR, above the UAV-only model's 38.89\% (\cref{tab:openfly-by-scene}) and the reduced-data settings (\cref{tab:full-scaling}).
Joint training continues to help on constant-altitude flight tasks.

\subsection{Separate and Joint Training}
\label{app:openfly-joint}

\Cref{tab:openfly-by-scene} compares the UAV-only model and NaviBridge by scene on the 1,795-task set with the reference altitude-change rule in \cref{app:openfly-protocol}.

\begin{table}[t]
    \caption{Separate and joint training on OpenFly by scene, with tasks involving reference altitude changes counted as failures. The overall scores weight tasks equally.}
    \label{tab:openfly-by-scene}
    \centering
    \small
    \renewcommand{\arraystretch}{1.15}
    \begin{tabular*}{.86\linewidth}{@{\extracolsep{\fill}}lrrrrr@{}}
        \toprule
        Scene & Tasks & \multicolumn{2}{c}{UAV only} & \multicolumn{2}{c}{NaviBridge} \\
        \cmidrule(lr){3-4}\cmidrule(l){5-6}
              & & SR (\%) & SPL (\%) & SR (\%) & SPL (\%) \\
        \midrule
        16 & 201 & 26.37 & 11.27 & 40.30 & 37.66 \\
        18 & 203 & 41.87 & 19.48 & 61.58 & 57.47 \\
        23 & 200 & 39.00 & 18.42 & 44.00 & 39.77 \\
        26 & 202 & 35.15 & 13.92 & 56.93 & 54.71 \\
        sh & 202 & 40.10 & 18.67 & 56.93 & 51.02 \\
        gz & 201 & 21.39 & 16.95 & 38.31 & 35.38 \\
        bigcity & 182 & 55.49 & 27.40 & 63.74 & 49.05 \\
        smallcity & 404 & 46.04 & 37.82 & 47.28 & 42.59 \\
        \midrule
        Overall & 1,795 & 38.89 & 22.37 & 50.58 & 45.57 \\
        \bottomrule
    \end{tabular*}
\end{table}

\subsection{Goal-Region Entry and Termination}
\label{app:openfly-behavior}

The original paired analysis in \cref{sec:aerial-behavior} uses recorded rollouts from both models on 1,794 tasks with reference-altitude assistance.
The remaining task was excluded from this paired analysis because repeated rendering errors during evaluation prevented complete paired rollouts.
Including the remaining task could change either model's SR by less than 0.06 percentage points, preserving the observed improvement from joint training.
We record entry when the UAV has a 3D distance to the goal below 20\,m after the start.
The UAV-only and joint models enter the region on 1,350 and 1,398 tasks, respectively, and end outside it on 597 and 403 of those tasks.

On the 1,467 constant-altitude tasks, the UAV-only model enters the region on 1,139 tasks, finishing inside on 698 and outside on 441.
The joint model enters on 1,181, finishing inside on 908 and outside on 273.
The net gain of 210 successes consists of 42 more entries and 168 fewer failures after entry.
In the paired comparison, 363 failures become successes and 153 successes become failures.
Both models enter the region on 1,030 constant-altitude tasks, and successes on these tasks rise from 641 to 803.

The stopping-rate aggregation in \cref{sec:aerial-behavior} omitted BigCity.
The original seven-scene stopping analysis contains 1,317 constant-altitude tasks, including 900 on which both models enter the region; BigCity adds 150 constant-altitude tasks, including 130 with entry by both models.
For the stopping comparison, we include all eight scenes and exclude one paired task whose UAV-only rollout was interrupted by a rendering failure.
The constant-altitude task set therefore grows from 1,317 to 1,466, and the number with entry by both models grows from 900 to 1,029.
The 1,030 constant-altitude tasks with entry by both models reported above include this interrupted episode.

Among all 1,793 valid paired tasks, both models enter the region on 1,196 tasks.
Stops increase from 679 to 873, corresponding to stop rates of 56.77\% and 72.99\%.
Within the 1,029 constant-altitude tasks with entry by both models, stops increase from 638 to 803, or from 62.00\% to 78.04\%.
Across all eight scenes, stops at the first entry increase from 73 to 74; of the 194 net additional stops, 193 occur on decisions after the first entry.
These eight-scene results further support improved stopping near the goal under joint training.

\section{Real-World Evaluation}
\label{app:real-world}

\begin{figure*}[t]
    \centering
    \includegraphics[width=\textwidth]{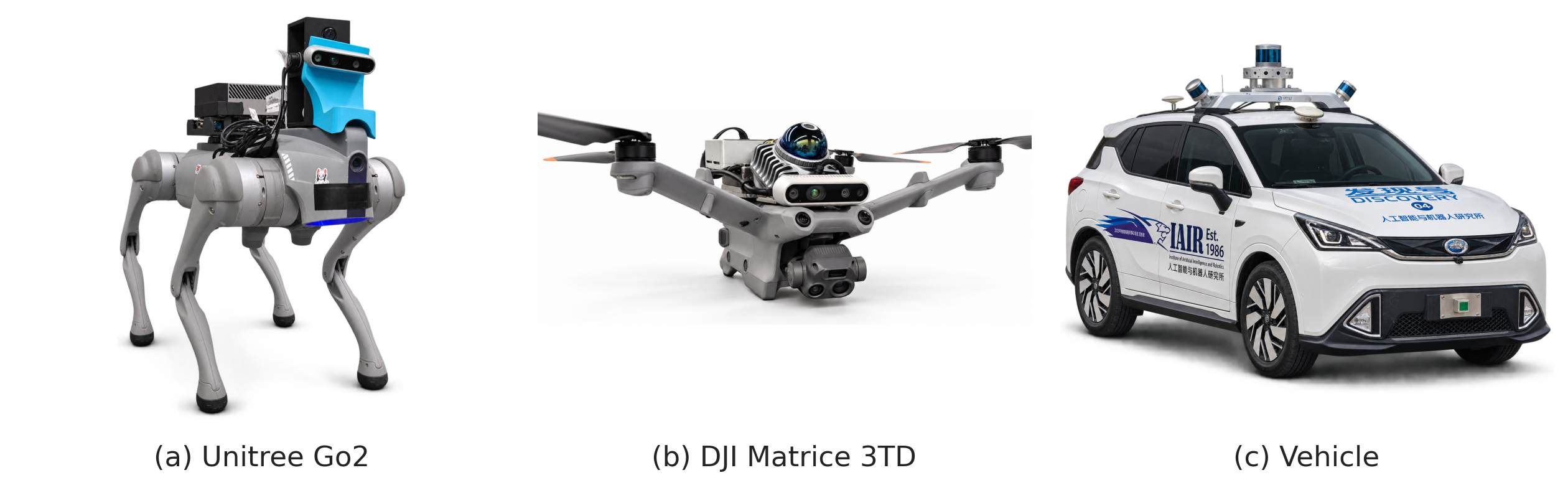}
    \caption{Quadruped, UAV, and vehicle platforms used in the real-world experiments.}
    \label{fig:supp-platforms}
\end{figure*}

\Cref{fig:supp-platforms} shows the three experimental platforms.
The quadruped is a Unitree Go2 equipped with a D435i camera and Jetson Orin.
The UAV is a DJI Matrice 3TD equipped with a RealSense D435 camera and MID-360 lidar, connected over a local network to a computer with an RTX 3070 GPU.
Its policy predicts horizontal waypoints from camera images and language instructions; lidar provides localization for trajectory visualization.
The vehicle records road scenes through a front-facing camera for open-loop waypoint prediction.

\subsection{Quadruped Point-Goal Navigation}

\begin{figure*}[t]
    \centering
    \includegraphics[width=\textwidth]{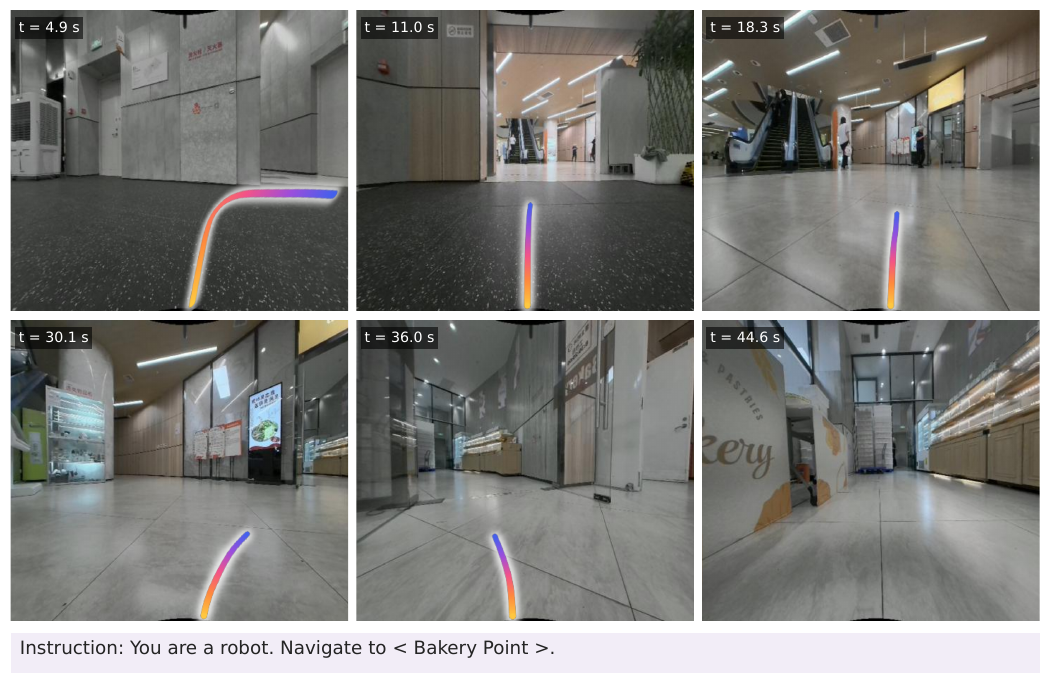}
    \caption{A continuous quadruped execution toward the goal, ordered by the original recording time. Colored curves show the waypoints predicted at each moment.}
    \label{fig:quadruped-extra}
\end{figure*}

The quadruped performs indoor point-goal navigation without real-world fine-tuning.
It generates waypoints with one denoising step, taking approximately 0.8 seconds per inference.
\Cref{fig:quadruped-extra} shows six moments from one continuous execution.
The robot starts facing a wall, turns into the corridor, passes the escalators and display area, follows the counter, and reaches the bakery.

\subsection{UAV Instruction Following}

\begin{figure*}[p]
    \centering
    \includegraphics[width=\textwidth,height=0.88\textheight,keepaspectratio]{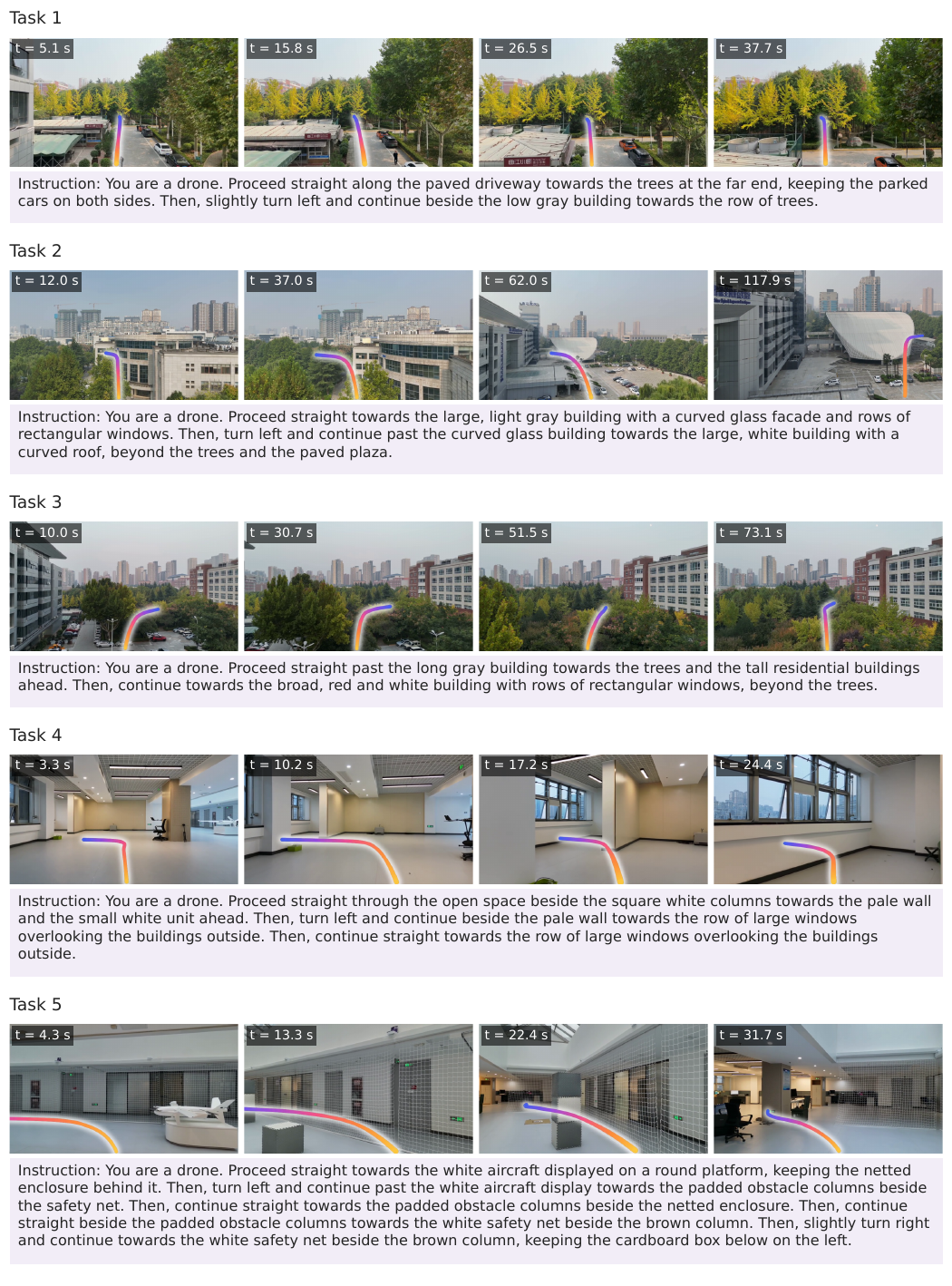}
    \caption{Continuous flight segments for Tasks~1--5. Each row belongs to one task and proceeds from left to right in video time. Colored curves show predicted waypoints.}
    \label{fig:uav-extra}
\end{figure*}

The UAV follows natural-language instructions in three outdoor and two indoor tasks.
It predicts five future horizontal waypoints with one denoising step, taking approximately one second per inference.
We collect one complete demonstration for each of four tasks and jointly fine-tune once on these demonstrations.
In \cref{fig:uav-extra}, Task~1 is a task-level zero-shot task without an adaptation demonstration, while Tasks~2--5 each have one demonstration.
The evaluation flight altitudes differ from the demonstration altitudes.

Tasks~1--3 involve following roads or buildings, turning, and approaching specified landmarks.
Tasks~4--5 involve traversing indoor open spaces, passing exhibits, and turning around columns.
Each row shows one task's complete instruction and execution segments.

\subsection{Open-Loop Prediction on Real Roads}

Local road regulations limit vehicle testing, so we evaluate waypoint prediction in a NAVSIM-style open-loop setting.
A front-facing in-car camera records 353.6 seconds of driving over approximately 754.2\,m.
NaviBridge takes the current image, four history images, and a navigation command as inputs and predicts waypoints with two denoising steps.
We generate the command from the recorded maneuver approximately three seconds later.
The inputs comprise 641 straight, 38 left-turn, and 18 right-turn prediction times.

\begin{figure*}[t]
    \centering
    \includegraphics[width=\textwidth]{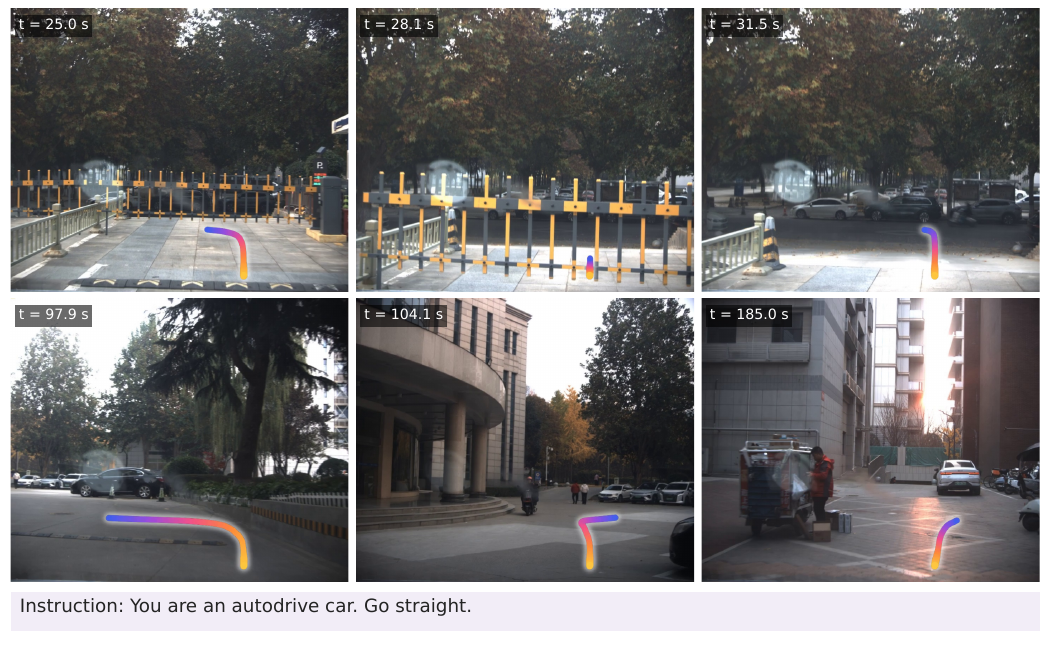}
    \caption{Open-loop predictions from the same real-road recording. The upper-left labels give video times, colored curves show the future waypoints predicted at each moment, and the bottom strip gives the navigation command.}
    \label{fig:driving-extra}
\end{figure*}

\Cref{fig:driving-extra} shows six moments from the road recording.
The first three frames cover the approach to a closed barrier and its subsequent opening.
The remaining frames cover a turn, passage near pedestrians, and a road between buildings.
The predicted curves are overlaid on the corresponding camera images, showing how predicted path length and direction change with the scene.

\section{Corrective versus Continued Training}
\label{app:correction-control}

To examine whether additional training time explains correction gains, we compare the common joint ground base model, models continued on base data alone, and models trained with correction data.
The base checkpoint at 12k steps gives the results used in the main paper.
Continued training retains the indoor and outdoor base data and their sampling proportions.
We evaluate checkpoints at 15k, 16k, and 21k steps, adding 3k, 4k, and 9k updates after 12k, respectively.
These durations cover the 2.75k, 3.5k, and 8.75k additional updates used by the three correction recipes.
We fix the three continued-training checkpoints before evaluation and compare all models on the same tasks and inference settings in \cref{app:ground-protocol}.

Continued training retains the base learning-rate schedule and optimizer state, whereas corrective training restarts optimization from the 12k weights with the lower learning rates and source proportions in \cref{tab:training-recipes}.
Their effective global batch sizes are 3,456 and 2,304, respectively.
The three continued-training runs therefore use approximately 10.37M, 13.82M, and 31.10M additional sample draws after 12k, exceeding the corresponding correction recipes' 6.34M, 8.06M, and 20.16M draws.

\begin{table}[!htbp]
    \caption{Continued training on base data versus corrective training. Updates are counted from the common 12k base checkpoint. Each row evaluates one checkpoint on all 240 indoor and 225 outdoor tasks. Bold denotes the best SR and SPL in each environment.}
    \label{tab:correction-control}
    \centering
    \small
    \setlength{\tabcolsep}{2.5pt}
    \begin{tabularx}{\columnwidth}{@{}Xrrr@{}}
        \toprule
        Recipe & Updates & Indoor & Outdoor \\
               & (k) & \multicolumn{2}{c}{SR / SPL (\%)} \\
        \midrule
        Joint ground base (12k) & 0 & 96.67 / 92.22 & 93.78 / 92.53 \\
        Continue base to 15k & 3 & 93.75 / 90.29 & 91.11 / 90.43 \\
        Continue base to 16k & 4 & 93.75 / 89.73 & 91.11 / 90.48 \\
        Continue base to 21k & 9 & 94.58 / 90.77 & 91.11 / 90.54 \\
        \midrule
        Indoor correction & 2.75 & \textbf{100.00 / 95.76} & 94.67 / 94.13 \\
        Outdoor correction & 3.5 & 95.83 / 94.33 & \textbf{95.11 / 94.99} \\
        Joint correction & 8.75 & \textbf{100.00} / 95.56 & 94.22 / 93.81 \\
        \bottomrule
    \end{tabularx}
\end{table}

All three continued-training models have lower indoor and outdoor SR and SPL than the 12k base model.
Their highest indoor SR is 94.58\%, and all three obtain 91.11\% outdoor SR.
With indoor or outdoor correction data, SR in the corresponding environment reaches 100.00\% and 95.11\%, respectively.
Longer training on base data does not reproduce these correction gains, supporting a shift in data collection toward policy deviations and recovery states once routine performance saturates.

\RemoveFromHook{env/tabular/begin}[navibridge]
\end{document}